%% file: neurips_2026.tex
\documentclass{article}

\PassOptionsToPackage{numbers, compress}{natbib}

\usepackage[main,final]{neurips_2026}

\usepackage[utf8]{inputenc} %
\usepackage[T1]{fontenc}    %
\usepackage{url}            %
\usepackage{booktabs}       %
\usepackage{amsfonts}       %
\usepackage{nicefrac}       %
\usepackage{microtype}      %
\usepackage{xcolor}         %

\usepackage{latexsym}
\usepackage{amsthm}
\usepackage{amsmath,mathtools}
\usepackage{amssymb}     %
\usepackage{multirow}       %
\usepackage{caption}        %
\usepackage[table]{xcolor}  %
\usepackage{arydshln}       %
\usepackage{wrapfig}
\usepackage{pifont}
\usepackage[T1]{fontenc}
\usepackage{fontawesome5}
\definecolor{kcgreen}{rgb}{0.1, 0.7, 0.2}

\usepackage{microtype}

\usepackage{inconsolata}

\usepackage{graphicx}
\usepackage{enumitem}
\usepackage[accsupp]{axessibility}  %

\usepackage{hyperref}
\newtheorem{theorem}{Theorem}  %
\newtheorem{proposition}{Proposition}

\theoremstyle{remark}               %

\usepackage{orcidlink}
\usepackage{marvosym}

\def\modelnm{ProCLIP}
\title{
\includegraphics[width=0.9cm]{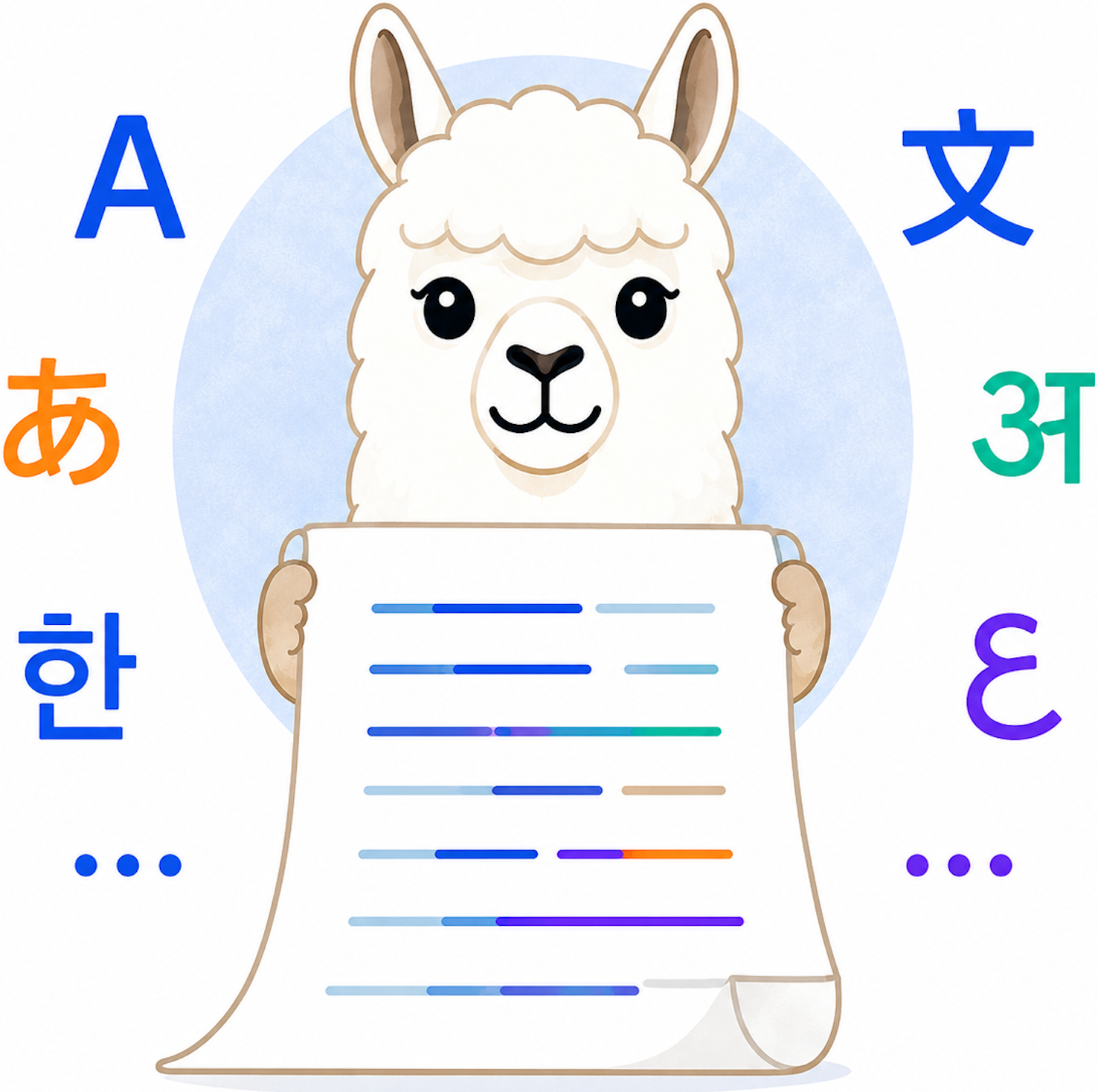}\hspace{-0.15em}
ProCLIP: Progressive Vision-Language 
Alignment via LLM-based Embedder
}

\newcommand{\plmark}{\textsuperscript{\dag}}

\author{%
\textbf{Xiaoxing Hu}\textsuperscript{1,2} \quad
\textbf{Kaicheng Yang}\textsuperscript{3}\plmark \quad
\textbf{Ziqi Ye}\textsuperscript{4,2} \quad 
\textbf{Ziyang Gong}\textsuperscript{1}\quad 
\textbf{Qi Ming}\textbf{\textsuperscript{5}} \quad \\
\textbf{Zonghao Guo}\textbf{\textsuperscript{6}} \quad
\textbf{Yu Tian}\textbf{\textsuperscript{6}} \quad 
\textbf{Xiang An}\textbf{\textsuperscript{3}} \quad
\textbf{Ziyong Feng}\textbf{\textsuperscript{3}} \quad 
\textbf{Xue Yang}\textbf{\textsuperscript{1}\textsuperscript{\Letter}} \\[3pt]
\textsuperscript{1}SAIS \& SCS, Shanghai Jiao Tong University \quad 
\textsuperscript{2}Shanghai Innovation Institute \\
\textsuperscript{3}DeepGlint \quad
\textsuperscript{4}Fudan University \quad
\textsuperscript{5}Beijing University of Technology \quad
\textsuperscript{6}Tsinghua University \\[3pt]
\plmark Project lead \quad \textsuperscript{\Letter}Corresponding author \\[3pt]
\faGithub\ Code: \href{https://github.com/VisionXLab/ProCLIP}{https://github.com/VisionXLab/ProCLIP}
}

\begin{document}

\maketitle

\input{Section/abstract}
\input{Section/introduction}
\input{Section/related_work}

\input{Section/methodology}
\input{Section/experiments}
\input{Section/conclusion}

\begin{ack}
This work was supported by the National Natural Science Foundation of China (62506229), Natural Science Foundation of Shanghai under 25ZR1402268, and Shanghai QiYuan Innovation Foundation.
\end{ack}

\bibliographystyle{plainnat}
\bibliography{main}

\newpage
\input{checklist.tex}

\clearpage
\appendix
\input{Section/appendix}

\end{document}

%% file: Section/abstract.tex
\begin{abstract}
Contrastive Language-Image Pre-training (CLIP) is fundamentally limited by its 77-token restriction, lack of multilingual capabilities, and coarse-grained semantic representations. While replacing CLIP’s native text encoder with a Large Language Model (LLM)-based embedder offers a promising solution, direct, from-scratch alignment can disrupt the semantic structures learned during pre-training. This often leads to degradation of cross-modal knowledge and particularly impairs recognition capabilities in zero-shot scenarios. In this paper, we propose \textbf{ProCLIP}, a curriculum-learning-inspired progressive alignment framework designed to systematically bridge the LLM embedder and CLIP's visual space, unlocking CLIP's potential for long-text, multilingual, and fine-grained understanding. Our framework operates in two stages: (1) \textbf{Representation Inheritance}, which distills CLIP's original text-space knowledge into an LLM adapter to establish a robust initial vision-language prior, and (2) \textbf{Contrastive Tuning}, which refines the cross-modal alignment with self-distillation regularization on the image encoder to further prevent catastrophic forgetting. To maintain semantic and geometric consistency, we introduce instance-level semantic and global structural alignment constraints throughout both stages. Extensive experiments demonstrate that ProCLIP improves zero-shot classification accuracy by 6.8\%--13.5\% over existing LLM-augmented baselines and achieves state-of-the-art performance in diverse long-text, multilingual, and fine-grained cross-modal retrieval tasks under comparable settings. 
\end{abstract}

%% file: Section/introduction.tex
\section{Introduction}
CLIP~\cite{clip} has demonstrated strong vision-language representations via large-scale contrastive learning, serving as a foundation for diverse downstream tasks, including retrieval~\cite{clipgen}, generation~\cite{alip}, and detection~\cite{cora}. Despite its success, CLIP's native text encoder is constrained by a strict 77-token limit and primarily relies on English-only supervision~\cite{longclip, siglip2}. Moreover, the lack of fine-grained textual supervision limits its ability to capture nuanced semantic details, restricting applicability in complex, real-world scenarios~\cite{degla}.

To overcome these limitations, recent works have sought to enhance CLIP's textual capabilities. Long-CLIP~\cite{longclip} interpolates positional embeddings to accommodate longer inputs, yet still struggles with fine-grained and multilingual understanding. Other methods, such as FLAME~\cite{flame} and LLM2CLIP~\cite{llm2clip}, replace CLIP's text encoder with Large Language Model (LLM)-based embedders~\cite{llm2vec, nvembedding} to leverage richer linguistic priors. However, these approaches typically employ a \emph{from-scratch alignment}, directly forcing the pre-trained image encoder to align with the LLM space via standard contrastive learning. By ignoring the valuable cross-modal knowledge embedded in CLIP, this naive strategy can disrupt pretrained semantic structures and degrade cross-modal representations, particularly affecting performance in zero-shot scenarios.

\begin{figure}[t!]
    \centering
    \includegraphics[width=1.0\linewidth]{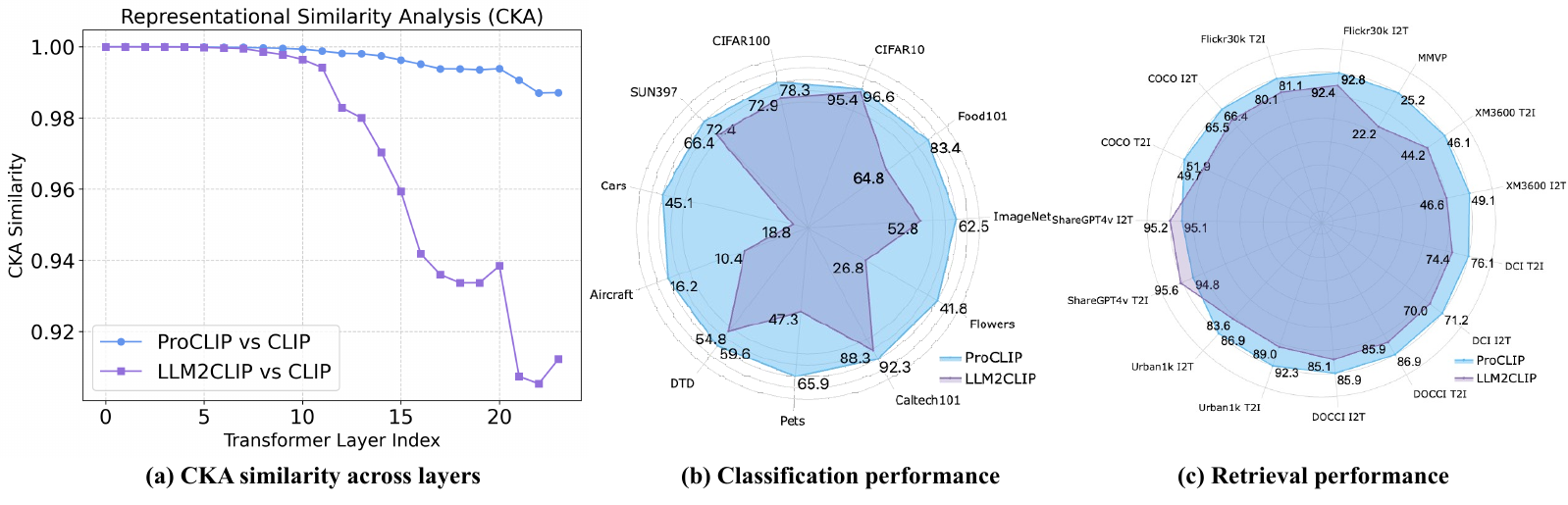}
    \vspace{-3mm}
    \caption{(a) CKA analysis of visual representations. (b) Comparison of classification performance and (c) retrieval performance between ProCLIP and our baseline LLM2CLIP. The figure uses a ViT-L base model trained on 3M samples. }
    \label{fig:cka}
    \vspace{-6mm}
\end{figure}

This motivates the following question: \textit{Can we leverage CLIP's pretrained knowledge to achieve effective cross-modal alignment with LLMs?} To address this challenge, we propose \textbf{ProCLIP}, a curriculum-learning-inspired progressive alignment framework that treats pretrained weights as foundational priors to bridge LLM embedders and CLIP's visual space. Rather than enforcing an abrupt semantic shift, ProCLIP adopts a two-stage curriculum:

\noindent(1) \textbf{Representation Inheritance}: We distill knowledge from the original CLIP text encoder into a lightweight adapter for the LLM. Instance-level semantic and global structural alignment constraints anchor the frozen LLM to CLIP's textual space, establishing a robust initial prior.  
\noindent(2) \textbf{Contrastive Tuning}: Building upon this foundation, we refine the representation space via contrastive tuning. This process preserves pretrained vision-language knowledge while improving cross-modal alignment.

We validate the preservation of pretrained knowledge via Centered Kernel Alignment (CKA) analysis on vision encoder features (Fig.~\ref{fig:cka} (a)). ProCLIP demonstrates superior retention of pre-trained knowledge compared to existing baselines, confirming that progressive alignment effectively facilitates cross-modal learning without sacrificing generalization. The framework offers multiple advantages: 1) \emph{Simplicity}: an intuitive, easy-to-implement design using pretrained weights as a semantic bridge; 2) \emph{Superiority}: state-of-the-art performance under identical data constraints (Fig.~\ref{fig:cka} (b) and (c)); 3) \emph{Versatility}: supports long-text, multilingual, and fine-grained scenarios; 4) \emph{Potential}: a robust paradigm for modular upgrades in multi-encoder foundation models.

Our main contributions are summarized as follows:
\begin{itemize}
    \item We \textbf{identify a critical limitation} in current LLM-augmented CLIP models: the from-scratch alignment neglects pretrained priors, leading to degraded cross-modal representations and impaired zero-shot performance.
    \item We \textbf{propose ProCLIP}, a curriculum-learning-inspired progressive alignment framework that effectively leverages pretrained knowledge to gradually build robust cross-modal representations, substantially preserving zero-shot recognition capabilities.
    \item We \textbf{demonstrate ProCLIP's efficacy} through extensive experiments. Under identical training constraints, ProCLIP achieves 6.8\%--13.5\% absolute gains in zero-shot classification and shows superior performance across diverse short/long-text, multilingual, and fine-grained cross-modal retrieval tasks.
\end{itemize}

%% file: Section/related_work.tex
\section{Related Work}
\noindent{\textbf{Vision-Language Contrastive Learning.}} Vision-language contrastive learning, exemplified by CLIP~\cite{clip}, aims to align multimodal representations in a shared semantic space via large-scale pretraining. As a foundational bridge between modalities, CLIP enables diverse open-vocabulary tasks, including image classification~\cite{CoOp,CoCoOp,aapl}, semantic segmentation~\cite{maskclip_seg,lseg,openseg,baseline,catseg,clearclip}, and object detection~\cite{detpro,mm_ovod}. Despite its versatility, CLIP is fundamentally constrained by its text encoder’s fixed input length and limited capacity, which hampers its performance in processing long-form, multilingual, and fine-grained semantics. To address these bottlenecks, Long-CLIP~\cite{longclip} extends input lengths through positional embedding interpolation, while LoTLIP~\cite{lotlip} aggregates diverse textual information using corner tokens to enhance long-text understanding. However, these methods remain tethered to the original CLIP text encoder's architecture, preventing the integration of broader open-world knowledge and leaving multilingual support unresolved.

\noindent{\textbf{LLM-based Representation Learning.}} Large Language Models (LLMs) have demonstrated exceptional proficiency across diverse natural language processing tasks~\cite{llama,gpt4,qwen,deepseekv3}. Recent research has pivoted toward repurposing decoder-only architectures for robust representation learning. Notably, LLM2Vec~\cite{llm2vec} transforms LLMs into versatile encoders via bidirectional attention and contrastive alignment, while Qwen3-Embedding~\cite{qwen3emb} leverages massive-scale pretraining to achieve state-of-the-art performance on the MTEB benchmark~\cite{mteb}. Inspired by these advancements, recent works~\cite{llm2clip, flame, sail} replace CLIP’s original text encoder with LLM-based embedders to handle multilingual, long-form, and complex textual inputs. However, these methods often employ coarse alignment strategies that inadvertently degrade the model's inherent generalization. Addressing this gap by developing more refined and efficient alignment techniques remains a critical and open challenge.

%% file: Section/methodology.tex
\begin{figure*}[t!]
    \centering
    \includegraphics[width=0.9\linewidth]{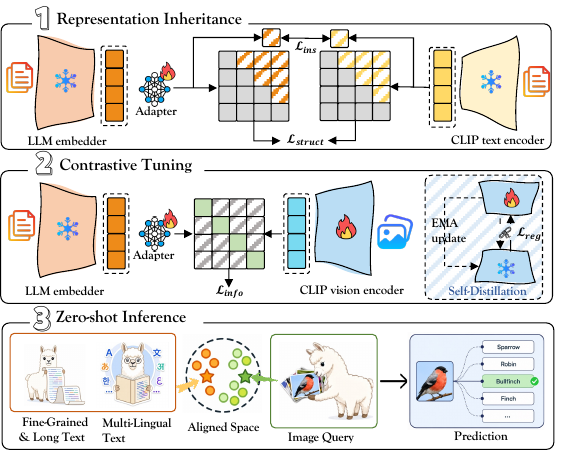}
    \vspace{-2mm}
    \caption{The pipeline of our proposed ProCLIP. It consists of representation inheritance via cross-architecture distillation and contrastive tuning integrated with self-distillation regularization.}
    \label{fig:method}
    \vspace{-7mm}
\end{figure*}
\section{Methodology}

In this section, we first introduce the preliminary concepts~(Sec.~\ref{preliminary}), including contrastive language-image pre-training and improving CLIP with an LLM-based embedder. We then present our proposed \textbf{ProCLIP} framework, which comprises two primary stages: 1) Representation Inheritance via Cross-Architecture Distillation (Sec.~\ref{sec:stage1}) and 2) Contrastive Tuning with Self-Distillation Regularization (Sec.~\ref{sec:stage2}). For a more detailed theoretical analysis, please refer to the Appendix.

\subsection{Preliminary}
\label{preliminary}

\noindent\textbf{Contrastive Language-Image Pre-training.} 
CLIP~\cite{clip} learns joint embeddings for images and text from large-scale image-text pairs using contrastive learning. Given a batch of image-text pairs $\{(I_i, T_i)\}_{i=1}^{\mathcal{B}}$, the pre-trained image encoder $\mathcal{E}_I$ and text encoder $\mathcal{E}_T$ map them into joint semantic representations $\{(v_i, t_i)\}_{i=1}^{\mathcal{B}}$. The dual-tower architecture is optimized with a symmetric InfoNCE loss:
\begin{equation}
\label{eq:contrast}
\mathcal{L}_{\text{CLIP}} = -\sum_{i=1}^{\mathcal{B}} \left[ \log \frac{\exp(v_i \cdot t_i / \tau)}{\sum_{j=1}^{\mathcal{B}} \exp(v_i \cdot t_j / \tau)} + \log \frac{\exp(t_i \cdot v_i /\tau)}{\sum_{j=1}^{\mathcal{B}}\exp(t_i\cdot v_{j} /\tau)} \right],
\end{equation}
where $\tau$ is a learnable temperature. However, the native CLIP text encoder $\mathcal{E}_T$ is limited to sequences of 77 tokens.

\noindent\textbf{Improving CLIP with an LLM-based Embedder.}
Recent approaches such as LLM2CLIP~\cite{llm2clip} replace the CLIP text encoder with a pre-trained LLM-based embedder $\mathcal{G}_T$ to improve long-text comprehension, multilingual understanding, and fine-grained semantic discrimination. Text inputs $\{T_i\}$ are encoded offline as embeddings $\{t'_i = \mathcal{G}_T(T_i)\}$. During fine-tuning, an MLP-based adapter projects these embeddings into the CLIP visual space.

Directly applying contrastive learning on the LLM embeddings and visual encoder is challenging because the LLM embedding space initially lacks alignment with the visual encoder. Unconstrained fine-tuning on limited data can lead to drift in the image encoder, reducing zero-shot generalization. ProCLIP addresses this with a progressive, two-stage alignment strategy.

\subsection{Stage I: Representation Inheritance via Cross-Architecture Distillation}
\label{sec:stage1}

Given pre-trained CLIP encoders $\{\mathcal{E}_I, \mathcal{E}_T\}$ and a pre-trained LLM embedder $\mathcal{G}_T$, our goal is to replace $\mathcal{E}_T$ with $\mathcal{G}_T$ while preserving CLIP's established visual-textual relationships. We first extract textual embeddings offline: $t'_i = \mathcal{G}_T(T_i) \in \mathbb{R}^d$.

\noindent\textbf{Cross-Architecture Distillation.} 
We use the original CLIP text encoder $\mathcal{E}_T$ as a teacher to provide structural supervision to the LLM embeddings. This design leverages CLIP’s pretrained visual-textual space as a \textbf{stable structural scaffold}, rather than relying on CLIP to process long or multilingual text. The LLM embedder, which already encodes long-text and multilingual semantics, inherits these visual alignment priors, allowing it to produce embeddings that respect CLIP’s visual space while supporting long-text and multiple languages.

Specifically, we use an MLP adapter to project LLM embeddings to the CLIP dimension: $t_i^* = \text{Adapter}(t'_i)$, and define two alignment losses:  

\noindent\textbf{Instance-level semantic alignment}:
    \begin{equation}
    \mathcal{L}_{\text{ins}} = \sum_{i=1}^{\mathcal{B}} \| t_i^* - e_i \|_2
    \end{equation}
    aligning individual LLM embeddings $t_i^*$ to the corresponding CLIP teacher embeddings $e_i = \mathcal{E}_T(T_i)$.
\noindent\textbf{Global structural alignment}:
    \begin{equation}
    \mathcal{L}_{\text{struct}} = \sum_{i<j}^{\mathcal{B}} | d(t_i^*, t_j^*) - d(e_i, e_j) |
    \end{equation}
    preserving inter-sample geometric relationships.
The overall Stage I loss is $\mathcal{L}_{\text{dis}} = \mathcal{L}_{\text{ins}} + \mathcal{L}_{\text{struct}}$.

\subsection{Stage II: Contrastive Tuning Integrated with Self-Distillation Regularization.}
\label{sec:stage2}

After the inheritance phase, the adapter has been preliminarily anchored to CLIP's established vision-language embedding space. This significantly eases the subsequent optimization landscape for cross-modal contrastive fine-tuning. We utilize the standard InfoNCE loss to fully align the image embeddings $v_i = \mathcal{E}_I(I_i)$ with the projected LLM embeddings $t_i^*$:
\begin{equation}
\label{eq:info}
\mathcal{L}_{\text{info}} = -\sum_{i=1}^{\mathcal{B}}\left[ \log \frac{\exp(v_i \cdot t_i^{*} / \tau)}{\sum_{j=1}^{\mathcal{B}} \exp(v_i \cdot t_j^{*} / \tau)} + \log \frac{\exp(t_i^* \cdot v_i /\tau)}{\sum_{j=1}^{\mathcal{B}}\exp(t_i^*\cdot v_j /\tau)} \right].
\end{equation}

Beyond standard contrastive learning, it is imperative to mitigate the catastrophic forgetting of pre-trained visual knowledge during this adaptation. Therefore, we impose a \textbf{self-distillation regularization constraint} on the CLIP image encoder. Symmetrical to the knowledge transfer in Stage I, we apply both instance-level and structural constraints on the visual side. Let $v_i^* = \mathcal{E}^*_I(I_i)$ and $v_j^* = \mathcal{E}^*_I(I_j)$ denote the ``teacher'' vision embeddings derived from an EMA-updated image encoder. The regularization loss $\mathcal{L}_{\text{reg}}$ is formulated as:
\begin{equation}
\label{eq:reg}
\mathcal{L}_{\text{reg}} = \sum_{i=1}^{\mathcal{B}} \| v_i^* - v_i \|_2 + \sum_{i<j}^{\mathcal{B}} | d(v_i^*, v_j^*) - d(v_i, v_j) |.
\end{equation}
The teacher model $\mathcal{E}_I^*$ is updated via EMA from the student image encoder $\mathcal{E}_I$ at each training step to provide stable, historical targets:
\begin{equation}
\label{eq:ema}
\mathcal{E}_I^* \leftarrow \alpha \mathcal{E}_I^* + (1-\alpha)\mathcal{E}_I,
\end{equation}
where $\alpha \in [0, 1)$ controls the momentum update rate. The overall objective function is thus defined as:
\begin{equation}
\mathcal{L}_{\text{tune}} = \mathcal{L}_{\text{info}} + \lambda_{\text{reg}}\mathcal{L}_{\text{reg}},
\end{equation}
where $\lambda_{\text{reg}}$ is a hyperparameter balancing the contrastive alignment and the self-distillation penalty.

%% file: Section/experiments.tex
\section{Experiments}\label{sec:exp}
\input{Table/retrieval}
\begin{figure*}[t!]
\centering
\includegraphics[width=\linewidth]{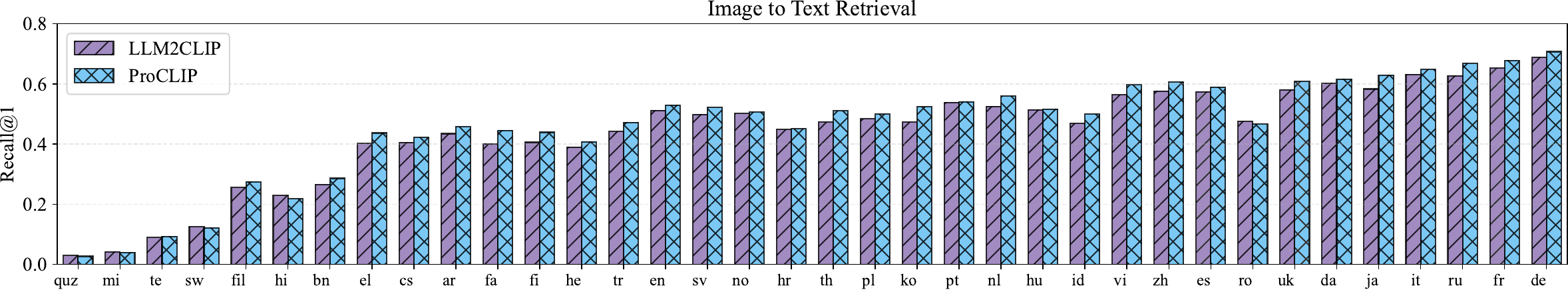} \\
\vspace{1mm} %
\includegraphics[width=\linewidth]{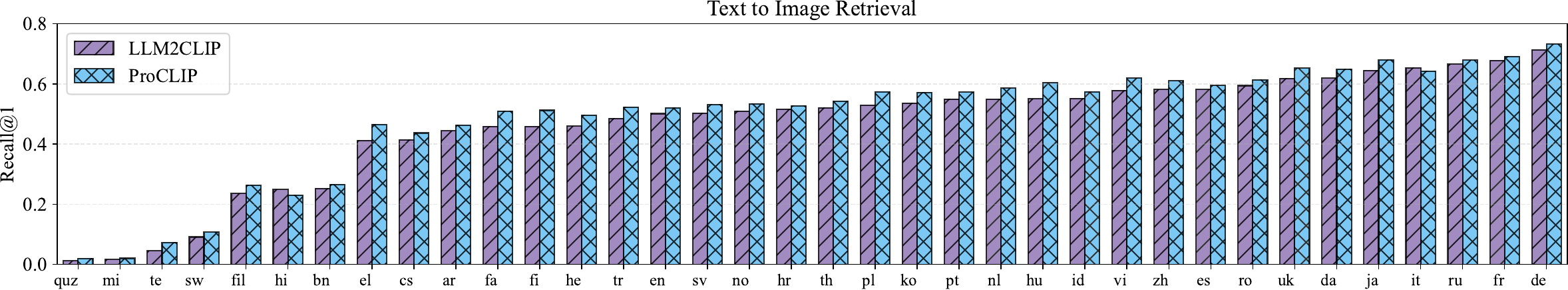}
\vspace{-5mm}
\caption{Per-language image-text retrieval performance on the XM3600 benchmark.}
\label{fig:xm3600}
\vspace{-5mm}
\end{figure*}

\subsection{Experimental Setup}

\noindent{\textbf{Datasets and Benchmarks.}} 
For the alignment pre-training phase, we utilize CC3M~\cite{cc12}, CC12M~\cite{cc12}, and YFCC15M~\cite{yfcc100m}, combined with curated high-quality captions from DreamLIP~\cite{dreamlip}. To systematically explore the effects of data scaling, we conduct experiments across three data volumes: 3M (CC3M), 15M (CC3M + CC12M), and 30M (CC3M + CC12M + YFCC15M). 
For downstream evaluation, we comprehensively assess the model across multiple dimensions: (1) zero-shot classification on 11 diverse datasets, (2) robustness evaluations on 5 ImageNet variants, (3) standard cross-modal retrieval on 6 benchmarks, (4) multilingual retrieval on XM3600~\cite{thapliyal2022crossmodal}, and (5) fine-grained compositional understanding on SugarCrepe~\cite{sugarcrepe} and MMVP-VLM~\cite{mmvp}. Additional experiments and evaluations are provided in the Appendix.\input{Table/zero_shot}
\begin{figure}[t!]
    \centering
    \begin{minipage}[t]{0.48\linewidth}
        \centering
        \vspace{0pt} %
        \includegraphics[width=\linewidth]{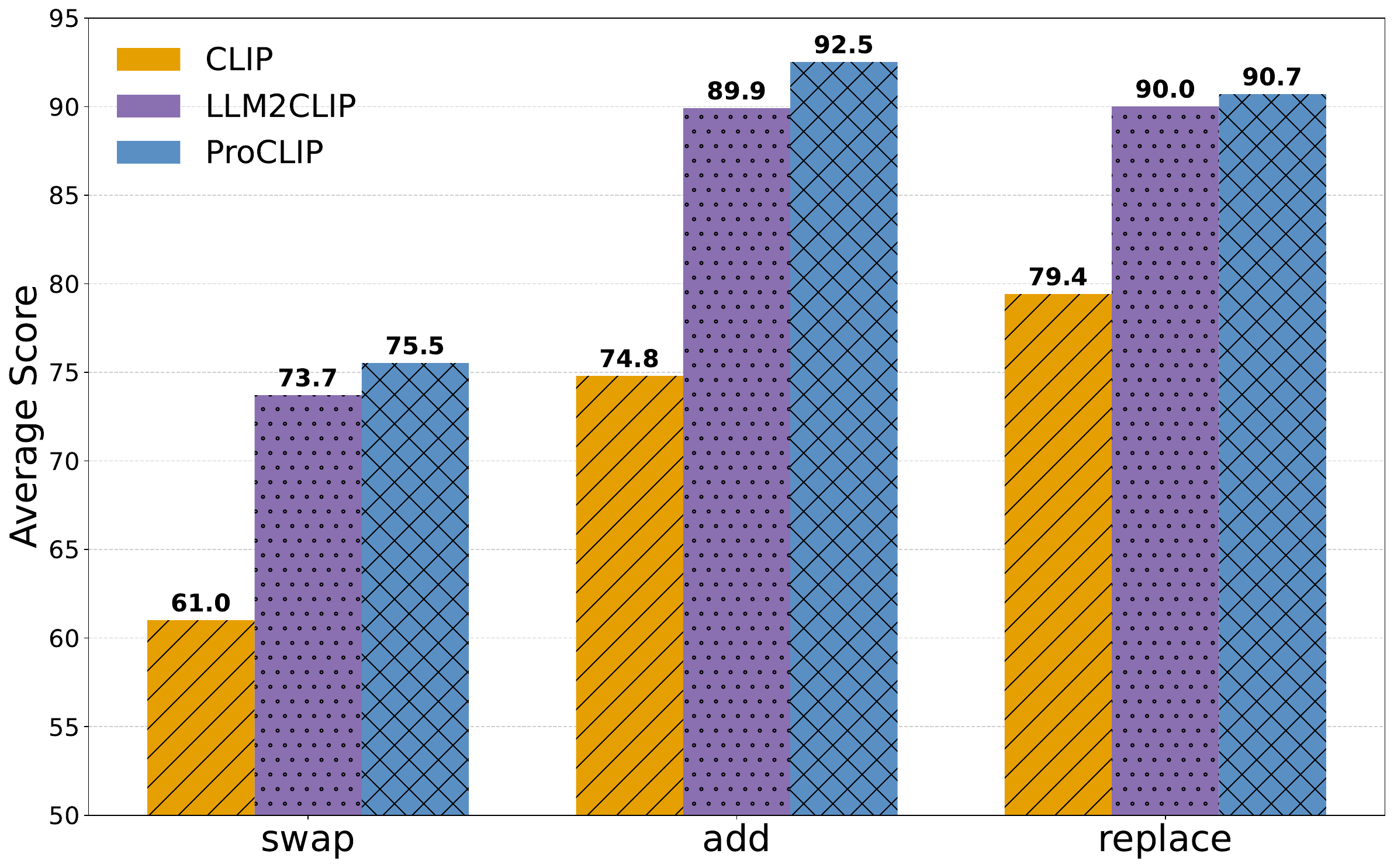}
        \caption{Performance on SugarCrepe.}
        \label{fig:sugarcrepe}
    \end{minipage}
        \hfill %
    \begin{minipage}[t]{0.48\linewidth}
        \centering
        
        \vspace{0pt} %
        \captionsetup{type=table}
        \caption{Robustness performance. The best results are marked in \textbf{bold}.}
        \label{tab:robustness_main}
        \resizebox{1.0\linewidth}{!}{
            \begin{tabular}{lcccccccc}
                \toprule
                \multirow{2}{*}{\textbf{Method}} & \multirow{2}{*}{\textbf{\#Data}} & \multicolumn{5}{c}{\textbf{Robustness}} \\
                & & \textbf{IN-V2} & \textbf{IN-A} & \textbf{IN-O} & \textbf{IN-R} & \textbf{IN-S} & \textbf{Avg.} \\
                \midrule
                \multicolumn{4}{l}{\textit{Model Architecture: CLIP ViT-L/14}} & & \\
                \textcolor{gray}{CLIP} & \textcolor{gray}{400M} & \textcolor{gray}{69.8} & \textcolor{gray}{70.8} & \textcolor{gray}{32.2} & \textcolor{gray}{87.8} & \textcolor{gray}{59.6}  & 64.0\\
                LLM2CLIP & 3M & 49.0 & 46.6 & \textbf{32.4} & 75.0 & 44.8 & 49.5\\
                \rowcolor[rgb]{0.925,0.957,1}\modelnm\ & 3M & \textbf{58.3} & \textbf{63.3} & 31.6 & \textbf{84.0} & \textbf{52.3} &\textbf{57.9}\\
                \hdashline
                LLM2CLIP & 15M & 50.8 & 50.1 & 33.8 & 78.2 & 46.3 & 51.8 \\
                \rowcolor[rgb]{0.925,0.957,1}\modelnm\ & 15M & \textbf{62.1} & \textbf{66.4} & \textbf{34.2} & \textbf{86.4} & \textbf{55.3} &\textbf{60.9}\\
                \hdashline
                LLM2CLIP & 30M & 52.7 & 52.7 & 34.0 & 78.6 & 47.3  & 53.0\\
                \rowcolor[rgb]{0.925,0.957,1}\modelnm\ & 30M & \textbf{63.4} & \textbf{68.0} & \textbf{34.1} & \textbf{86.8} & \textbf{55.7} & \textbf{61.6}\\
                \bottomrule
            \end{tabular}
        }
    \end{minipage}
    \vspace{-5mm}
\end{figure}

\noindent{\textbf{Implementation Details.}}
For the LLM-based embedder, we primarily utilize LLaMA3-8B-CC, consistent with the LLM2CLIP baseline~\cite{llm2clip}.
During our progressive alignment, the Representation Inheritance phase is trained for four epochs, followed by another four epochs of Contrastive Tuning. See the Appendix for detailed parameter configurations and evaluation settings.

\subsection{Main Results}

\noindent{\textbf{Cross-Modal Retrieval.}} 
As shown in Tab.~\ref{tab:retrieval}, ProCLIP achieves stronger average retrieval performance than the competitive LLM2CLIP baseline across the evaluated model scales, while a few individual metrics remain comparable or slightly lower. In standard short-text scenarios (e.g., Flickr30k and COCO), ProCLIP yields substantial gains. Notably, using the ViT-L/14 architecture trained on 30M data, it achieves 94.5\% I2T R@1 on Flickr30k, surpassing LLM2CLIP by 1.4\%. ProCLIP also shows clear improvements on several long-text benchmarks (e.g., DCI and Urban-1k). For instance, on the DCI dataset, ProCLIP (ViT-B/16, 30M) attains remarkable scores of 73.6\% I2T and 76.9\% T2I R@1. These results validate ProCLIP's versatility in handling both concise and complex textual descriptions.
\begin{figure}[!t]
\centering
\begin{minipage}[t]{0.35\linewidth}
\centering
\vfill
\includegraphics[width=\linewidth]{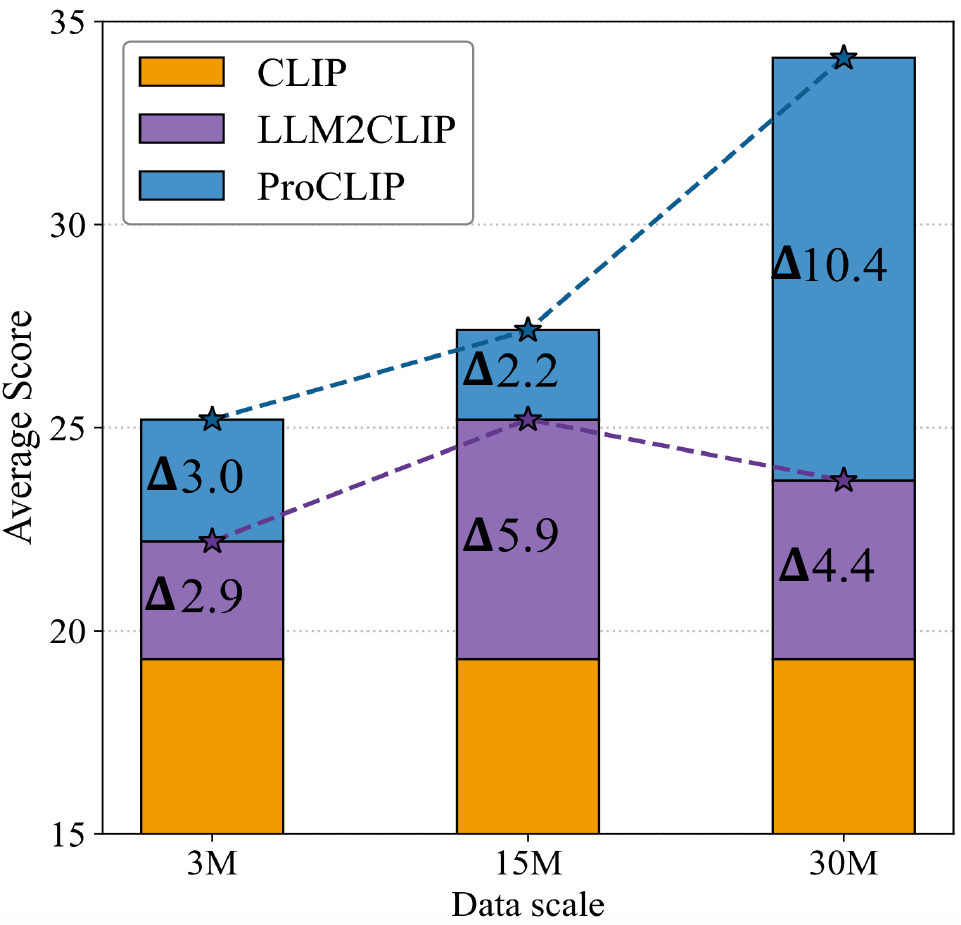}
\vspace{-4mm}
\caption{MMVP performance.}
\vspace{-6mm}
\label{fig:mmvp}
\end{minipage}
\begin{minipage}[t]{0.62\linewidth}
\centering
\captionsetup{type=table}
\caption{Comparison with other methods across different model scales and LLM embedders.}

\resizebox{\linewidth}{!}{%
\setlength{\tabcolsep}{2.5pt}
\begin{tabular}{lllcccccc} %
\toprule
\multirow{2}{*}{\textbf{Method}} &
\multirow{2}{*}{\textbf{ViT Init}} &
\multirow{2}{*}{\textbf{LLM Embedder}} & 
\multirow{2}{*}{\textbf{\#Data}} &
\multirow{2}{*}{\textbf{IN-1k}} &
\multicolumn{2}{c}{\textbf{COCO}} &
\multicolumn{2}{c}{\textbf{Flickr30k}} \\
& & & &  & 
\textbf{I2T} & \textbf{T2I} &
\textbf{I2T} & \textbf{T2I} \\
\midrule
FLAME & random & Mistral-Nemo &  3M & 36.0 & 43.3 & 28.6 & 67.3 & 53.6\\
ShareLock & DINOv2 B/14  &   Llama3 &3M& 52.1 & - & -& - & -\\ 
LIFT & random & NV-Embedv2 & 512M & 43.6 & 34.6 & 36.0 & 69.1 & 72.9  \\

LiT &CLIP B/16 & Llama3-CC &  3M & 51.0 & 56.2 & 41.9 & 85.2 & 71.9 \\

LLM2CLIP &CLIP B/16 & Llama3-CC &  3M & 45.8 & 60.5 & 44.8 & 88.0 & 75.3 \\
\rowcolor[rgb]{0.925,0.957,1}ProCLIP &CLIP B/16 & Llama3-CC &  3M & \textbf{54.8} & \textbf{61.7} & \textbf{46.8} & \textbf{89.4} & \textbf{77.6} \\
\midrule
SAIL & DINOv2 L/14 & NV-Embedv2 & 3M & 54.0 & 45.4 & 32.9 & - & - \\ 
LiT &CLIP L/14 & Llama3-CC &  3M & 60.1 & 59.4 & 44.6 & 88.0 & 74.7   \\
LLM2CLIP &CLIP L/14 & Llama3-CC &  3M & 52.8 & 65.5 & 49.7 & 92.4 & 80.1   \\
\rowcolor[rgb]{0.925,0.957,1}ProCLIP &CLIP L/14& NV-Embedv2 &  3M & 61.4  & 64.8 & 51.7 & 91.9 & \textbf{81.4} \\
\rowcolor[rgb]{0.925,0.957,1}ProCLIP &CLIP L/14& Llama3-CC &  3M & \textbf{62.5} &  \textbf{66.4} & \textbf{51.9} & \textbf{92.8} & 81.1  \\
\bottomrule
\end{tabular}}
\label{tab:other}
\end{minipage}
\vspace{-5mm}
\end{figure}

\noindent{\textbf{Zero-Shot Classification.}} Tab.~\ref{tab:classification} summarizes the zero-shot classification results on 11 downstream tasks. Direct alignment methods like the baseline LLM2CLIP struggle to achieve optimal zero-shot generalization. In contrast, ProCLIP effectively addresses this issue, yielding consistent and significant absolute gains of 6.8\%--13.5\% over LLM2CLIP across all model architectures and data scales. Furthermore, when compared to the vanilla CLIP trained under the identical 30M data setting, ProCLIP demonstrates substantial superiority (e.g., achieving 59.0\% vs. 48.9\% on ViT-B/32, and 63.0\% vs. 51.4\% on ViT-B/16). This benefit mainly comes from our curriculum alignment strategy, where self-distillation mitigates representation collapse while maintaining robust cross-modal alignment. ProCLIP still lags behind Long-CLIP and FG-CLIP in zero-shot classification, though it is worth noting that FG-CLIP relies on a massively larger 1.6B dataset. Nevertheless, ProCLIP, similar to LLM2CLIP, aims at a broader CLIP extension, improving not only cross-modal retrieval but also multilingual and fine-grained understanding. It therefore offers a more comprehensive trade-off across diverse cross-modal capabilities.

\noindent\textbf{Robustness and Multilingual Capabilities.} 
Tab.~\ref{tab:robustness_main} evaluates the robustness of our representations on the ViT-L/14 architecture. ProCLIP shows average gains of 8.4\%--9.1\% across different data scales in this setting. Most notably, on challenging out-of-distribution (OOD) benchmarks such as ImageNet-A and ImageNet-R, ProCLIP outperforms LLM2CLIP by over 8 percentage points. This highlights ProCLIP's superior capacity to handle distribution shifts and complex perturbations. 
Additionally, leveraging the semantic breadth of the LLM embedder, ProCLIP exhibits robust multilingual capabilities. As shown in Fig.~\ref{fig:xm3600}, ProCLIP consistently outperforms LLM2CLIP on the XM3600 benchmark~\cite{thapliyal2022crossmodal} across multiple languages (trained with 3M samples, using ViT-L base model). This superiority is primarily attributed to our progressive strategy, which effectively bridges the semantic gap between the original visual space and the vast, multilingual LLM space.

\input{Table/llava}

\noindent\textbf{Fine-Grained Understanding.} 
Fig.~\ref{fig:sugarcrepe} demonstrates the significant potential of ProCLIP in complex compositional reasoning tasks, where all models are trained on 3M samples using a ViT-L base model. It surpasses the original CLIP baseline by an average margin of 11.3\%--17.7\% on SugarCrepe. These improvements demonstrate that anchoring an LLM embedder to the visual space vastly enhances fine-grained semantic discrimination. Fig.~\ref{fig:mmvp} further presents the fine-grained understanding performance on the challenging MMVP benchmark~\cite{mmvp}, trained with ViT-L base model. While LLM2CLIP improves over CLIP by 2.9\%--5.9\%, ProCLIP dramatically advances these results, achieving additional gains of up to 10.4\% relative to the baseline.

\noindent\textbf{Comparison with other LLM-augmented  Methods.} To rigorously prove the effectiveness of ProCLIP, we provide a comprehensive comparison with recent state-of-the-art LLM-augmented CLIP models, including FLAME~\cite{flame}, ShareLock~\cite{sharelock}, LIFT~\cite{lift}, SAIL~\cite{sail}, LiT~\cite{lit}, and LLM2CLIP~\cite{llm2clip}. As detailed in Tab.~\ref{tab:other}, under identical or lower training budgets, ProCLIP consistently achieves superior performance. Benefiting from representation inheritance and self-distillation, ProCLIP not only excels in ImageNet-1k classification but also significantly enhances general retrieval capabilities.

\noindent\textbf{MLLM Performance.}
As shown in Tab.~\ref{tab:llava}, when integrating the fine-tuned vision encoder into the MLLM (using LLaVA1.5-7B~\cite{llava_1.5} framework), we observe performance improvements over CLIP on most benchmarks. This can be mainly attributed to the alignment with high-quality data, which enhances the semantic representation capability of the vision encoder. This also validates the flexibility of the ProCLIP paradigm, which supports both cross-modal retrieval via a dual-encoder architecture and the utilization of standalone vision encoder representations for downstream tasks.

\begin{figure}[t!]
\centering
\begin{minipage}[t]{0.4\linewidth}
\centering
\vspace{0pt}
\includegraphics[width=\linewidth]{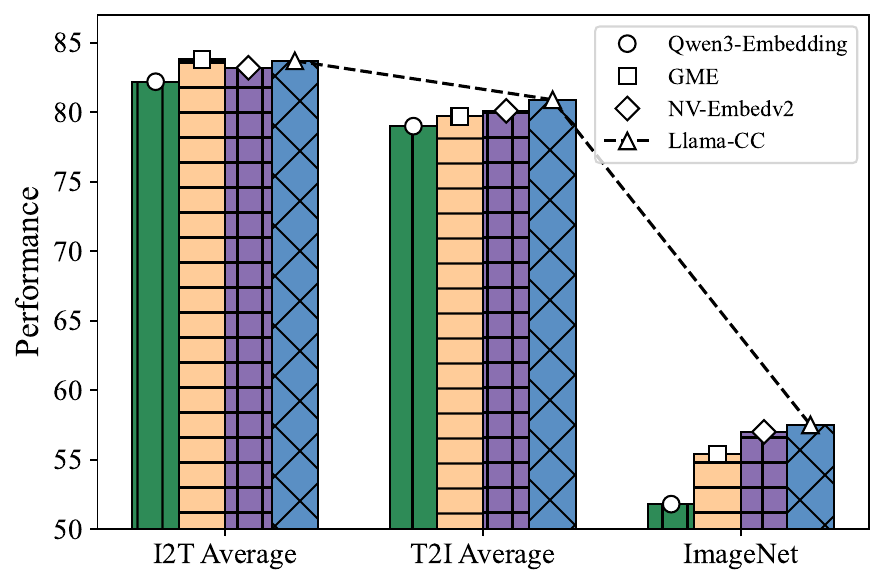}

\caption{Ablation on LLM embedders.}
\label{fig:embedder}
\end{minipage}
\vspace{-2mm}
\begin{minipage}[t]{0.57\linewidth}
\vspace{0pt}
\centering
\captionsetup{type=table}
\caption{Ablation on different components.}
\resizebox{\linewidth}{!}{%
\setlength{\tabcolsep}{2pt}
\begin{tabular}{l cccc ccc} %
\toprule
\multirow{2}{*}{\textbf{Method}}  & \multicolumn{2}{c}{\textbf{Stage 1}} & \multicolumn{2}{c}{\textbf{Stage 2}} & \multirow{2}{*}{\textbf{IN-1k}} & \multirow{2}{*}{\textbf{I2T Avg.}} &  \multirow{2}{*}{\textbf{T2I Avg.}} \\
&\textbf{$\mathcal{L}_{\text{ins}}$} & \textbf{$\mathcal{L}_{\text{struct}}$} & \textbf{$\mathcal{L}_{\text{info}}$} & \textbf{$\mathcal{L}_{\text{reg}}$}  &  &  & \\
\midrule
\textcolor{gray}{CLIP} & & & & &   \textcolor{gray}{74.5} & \textcolor{gray}{66.9} &\textcolor{gray}{55.5} \\

\textcolor{gray}{LLM2CLIP} & & & & &  \textcolor{gray}{52.8} & \textcolor{gray}{82.0} & \textcolor{gray}{79.1}   \\
\midrule
\multirow{4}{*}{ProCLIP} & \ding{51} & & & & 58.9 & 69.3 & 59.4 \\
& \ding{51} & \ding{51} & & & 59.5 & 70.3 & 61.2 \\
& \ding{51} & \ding{51} & \ding{51} & & 59.2 & 82.9 & 80.2 \\
\rowcolor[rgb]{0.925,0.957,1}& \ding{51} & \ding{51} & \ding{51} & \ding{51} & \textbf{62.5} & \textbf{83.0} & \textbf{80.5} \\
\bottomrule
\end{tabular}
}
\label{tab:ablation}
\end{minipage}%
\vspace{-5mm}
\end{figure}

\subsection{Ablation and Analysis}
\textbf{Ablation of Different LLM-based Embedder.}
Fig.~\ref{fig:embedder} compares several LLM embedders (Qwen3~\cite{qwen3emb}, GME~\cite{gme}, NV-Embedv2~\cite{nvembedding}, and Llama3-CC~\cite{llm2clip}) paired with ViT-L/14 (trained with 15M samples). Among them, Llama3-CC achieves the best overall results in classification and retrieval. As for its superior classification performance, we hypothesize that this is primarily attributable to the incorporation of textual information from image-text pairs during its text embedding training phase.  This pre-adaptation to the text distribution in image-text pairs facilitates subsequent cross-modal alignment, enabling the model to fit the data distribution more effectively and converge more rapidly. Notably, while retrieval performance remains relatively consistent across embedders, ImageNet classification accuracy varies significantly. This suggests that the alignment discrepancy between each LLM's feature space and the original CLIP space differs, leading to varying degrees of generalization degradation after training.

\begin{table*}[t!] %
    \centering
    
    \begin{minipage}[t]{0.5\linewidth}
        \centering
        \caption{Ablation study on different base model architectures. All methods use 3M training samples.}
        \label{tab:ablation_arch}
        \setlength{\tabcolsep}{3pt}
        \resizebox{1.0\linewidth}{!}{
        \begin{tabular}{llccc}
            \toprule
            \textbf{Base Model} & \textbf{Method} & \textbf{IN-1k} & \textbf{I2T Avg.} & \textbf{T2I Avg.} \\
            \midrule
            \multirow{2}{*}{CLIP ViT-L}        
            & LLM2CLIP & 52.8 & 82.0 & 79.1 \\
            & \cellcolor[rgb]{0.925,0.957,1}ProCLIP  & \cellcolor[rgb]{0.925,0.957,1}\textbf{62.5} & \cellcolor[rgb]{0.925,0.957,1}\textbf{83.0} & \cellcolor[rgb]{0.925,0.957,1}\textbf{80.5} \\
            \midrule
            \multirow{2}{*}{EVA02-CLIP ViT-L}
            & LLM2CLIP & 56.4 & 83.6 & 80.9 \\
            & \cellcolor[rgb]{0.925,0.957,1}ProCLIP  & \cellcolor[rgb]{0.925,0.957,1}\textbf{66.8} & \cellcolor[rgb]{0.925,0.957,1}\textbf{84.2} & \cellcolor[rgb]{0.925,0.957,1}\textbf{82.2} \\
            \midrule
            \multirow{2}{*}{SigLIP2 SO/14@224}
            & LLM2CLIP & 63.8 & 84.6 & 81.9 \\
            & \cellcolor[rgb]{0.925,0.957,1}ProCLIP  & \cellcolor[rgb]{0.925,0.957,1}\textbf{67.7} & \cellcolor[rgb]{0.925,0.957,1}\textbf{86.3} & \cellcolor[rgb]{0.925,0.957,1}\textbf{83.9} \\
            \bottomrule
        \end{tabular}
        }
    \end{minipage}
    \hfill %
    \begin{minipage}[t]{0.38\linewidth}
        \centering
        \caption{Ablation of $\mathcal{L}_{reg}$.}
        \label{tab:reg_weight}
        \setlength{\tabcolsep}{3pt}
        \resizebox{1.0\linewidth}{!}{
        \begin{tabular}{lcccc}
        \toprule
        \textbf{$\lambda_{reg}$} & \textbf{IN-1k} & \textbf{I2T Avg.} & \textbf{T2I Avg.}\\
        \midrule
        0    & 59.2  & 82.9 & 80.2 \\
        \midrule 
        2e-4 & 60.3 & 82.8 & 80.3 \\
        \rowcolor[rgb]{0.925,0.957,1}
        4e-4  & 62.5 & \textbf{83.0} & \textbf{80.5} \\
        6e-4  & \textbf{62.6} & 82.6 & 79.8 \\
        1e-3 &  62.5 & 81.8 & 79.5\\
        \bottomrule
        \end{tabular}
        }
    \end{minipage}
    \vspace{-5mm}
\end{table*}
\begin{table}[t!]
    \centering
    \begin{minipage}[t]{0.48\textwidth}
        \centering
        \caption{Ablation on progressive training.}
        \label{tab:ab_multi_stage_training}
        \resizebox{\textwidth}{!}{
            \begin{tabular}{lcccc}
            \toprule
                \textbf{Method} & \textbf{IN-1k} & \textbf{I2T Avg.} & \textbf{T2I Avg.} & \textbf{Training Hours $\downarrow$} \\
            \midrule
                One stage & 58.4 & 82.6 & 79.8 & \textbf{1.5}\\
                \rowcolor[rgb]{0.925,0.957,1}
                Two stage & \textbf{62.5} & \textbf{83.0} & \textbf{80.5} & 1.7  \\
            \bottomrule
            \end{tabular}
        }
    \end{minipage}
    \hfill %
    \begin{minipage}[t]{0.48\textwidth}
        \centering
        \caption{Performance and training efficiency comparison.}
        \label{tab:performance_comparison}
        \resizebox{\textwidth}{!}{
            \begin{tabular}{l c c c c}
            \toprule
                \textbf{Method} & \textbf{IN-1k} & \textbf{I2T Avg.} & \textbf{T2I Avg.} & \textbf{Training Hours$\downarrow$} \\
            \midrule
                Adapter   & \textbf{62.5} & \textbf{83.0} & 80.5 & \textbf{1.7}  \\
                \rowcolor[rgb]{0.925,0.957,1}
                LoRA(r=16) & 61.8 & 82.6 & \textbf{80.6} & $\sim$15 \\
            \bottomrule
            \end{tabular}
        }
    \end{minipage}
    \vspace{-5mm}
\end{table}

\noindent\textbf{Ablation of Different Base Model.} Tab.~\ref{tab:ablation_arch} shows \modelnm\ excels across architectures, surpassing LLM2CLIP in classification and retrieval. The remaining classification gap relative to vanilla CLIP arises from the latter's overwhelming data scale, a dominant factor in zero-shot tasks.

\noindent\textbf{Ablation of $\lambda_{reg}$.} Tab.~\ref{tab:reg_weight} presents the sensitivity analysis of the regularization weight for $\lambda_{\text{reg}}$. The results indicate that 4e-4 offers an optimal balance and is thus selected as our final configuration. Furthermore, the empirical evidence suggests that excessively strong regularization hampers alignment, leading to suboptimal results.

\noindent\textbf{Ablation of Different Components.}
Tab.~\ref{tab:ablation} details the ablation study of our components. First, instance semantic distillation achieves 58.9\% ImageNet-1k zero-shot accuracy using text only, confirming the successful transfer of CLIP’s textual priors to the adapter. Adding structural alignment loss further boosts both classification and retrieval by enabling the LLM to capture CLIP’s global geometric structure beyond point-wise semantics. Subsequent image-text contrastive tuning significantly enhances retrieval but compromises classification due to image encoder overfitting. This is mitigated by self-distillation, which recovers classification accuracy from 59.2\% to 62.5\%. Ultimately, structured self-distillation yields the best results by stabilizing the image representation space and preserving pretrained knowledge, effectively balancing fine-tuning gains with generalization.

 \begin{wraptable}{r}{0.4\textwidth}
    \centering
    \vspace{-5mm}
    \caption{Ablation on adapter.}
    \resizebox{1.0\linewidth}{!}{
    \begin{tabular}{cccc}
    \toprule
    \textbf{Adapter} & \textbf{IN-1k} & \textbf{I2T Avg.} & \textbf{T2I Avg.}\\
    \midrule
    \rowcolor[rgb]{0.925,0.957,1}
    MLP & \textbf{62.5} & \textbf{83.0} & 80.5 \\
    GLU & 61.8 & 82.6 & \textbf{80.6}\\
    Transformer & 61.6 & 82.5 & \textbf{80.6} \\
    \bottomrule
    \end{tabular}
    }
    \label{tab:ab_adapter_structure}
    \vspace{-5mm}
\end{wraptable}

\noindent\textbf{One stage vs Multi stage.} As shown in Tab.~\ref{tab:ab_multi_stage_training}, collapsing our two-stage strategy into a single stage significantly degrades both generalization and retrieval metrics. This confirms that our curriculum learning-inspired approach is essential for achieving superior cross-modal alignment.

\noindent\textbf{Ablation of adapter selection.}
To ensure a fair evaluation of ProCLIP, we maintain baseline consistency in adapter selection and fix all configurations at ~67M parameters. Our ablation studies  compare Transformer (1-layer), MLP (4-layer), and GLU (4-layer) adapters. Results in Tab.~\ref{tab:ab_adapter_structure} indicate that the MLP architecture offers the optimal balance between computational efficiency and accuracy.

\noindent\textbf{Offline Training vs. LoRA Fine-tuning.}
LLM embedders can be trained using PEFT methods (e.g., LoRA) or by extracting features offline and training only the adapter. As shown in Tab.~\ref{tab:performance_comparison}, using LoRA requires active LLM computation during training, which incurs substantial computational overhead without providing additional performance gains. Therefore, in our final setup, we train only the adapter following the LLM embedder, keeping the setting consistent with the baseline.

%% file: Table/retrieval.tex
\begin{table*}[t!]
\centering
\caption{Cross-modal retrieval performance Recall@1 on multiple datasets. \textbf{Bold values} indicate the better performance between LLM2CLIP and ProCLIP.}

\resizebox{\textwidth}{!}{
\begin{tabular}{lccccccccccccccc}%
\toprule
\multirow{2}{*}{\textbf{Method}} & \multirow{2}{*}{\textbf{\#Data}} & 
\multicolumn{2}{c}{\textbf{Flickr30k}} & \multicolumn{2}{c}{\textbf{COCO}} &
\multicolumn{2}{c}{\textbf{ShareGPT4V}} & \multicolumn{2}{c}{\textbf{Urban-1k}} &
\multicolumn{2}{c}{\textbf{DOCCI}} & \multicolumn{2}{c}{\textbf{DCI}} & 
\multicolumn{2}{c}{\textbf{Avg.}} \\
& & I2T & T2I & I2T & T2I & I2T & T2I & I2T & T2I & I2T & T2I & I2T & T2I & I2T & T2I \\
\midrule
\multicolumn{6}{l}{\textit{Model Architecture: CLIP ViT-B/32}} & & & &  & & & & & \\

LLM2CLIP & 3M & 83.5 & 70.1 & 55.6 & 41.1 & 94.2 &  \textbf{93.4} & 78.2 & 84.2 & 76.2 & 77.1 & 62.2 & 64.4 & 75.0 & 71.7 \\
\rowcolor[rgb]{0.925,0.957,1}\modelnm\ & 3M & \textbf{86.0} &  \textbf{73.5} &  \textbf{57.8} &  \textbf{43.5} &  \textbf{94.4} & 92.6 &  \textbf{80.8} &  \textbf{85.3} &  \textbf{78.1} &  \textbf{79.5} &  \textbf{65.7} &  \textbf{68.3} &  \textbf{77.1(\textcolor{kcgreen}{+2.1})} &  \textbf{73.8(\textcolor{kcgreen}{+2.1})} \\
\hdashline

LLM2CLIP & 15M & 86.2 & 72.2 & 58.5 & 43.2 &  \textbf{95.3} &  \textbf{94.2} & 80.6 &  \textbf{85.3} &  \textbf{79.2} &  \textbf{80.7} & 64.3 & 67.6 & 77.4 & 73.9 \\

\rowcolor[rgb]{0.925,0.957,1}\modelnm\ & 15M & \textbf{86.6} & \textbf{72.6} &  \textbf{59.0} &  \textbf{43.5} & 94.5 & 93.9 &  \textbf{82.2} &  \textbf{85.3} & 78.4 & 80.6 & \textbf{67.1} & \textbf{69.2} &  \textbf{78.0(\textcolor{kcgreen}{+0.6})} &  \textbf{74.2(\textcolor{kcgreen}{+0.3})} \\
\hdashline

CLIP & 30M & 85.6 & 69.5 & 59.3 & 42.8 
&76.8 & 67.3 &62.5 & 43.2 & 57.2 & 54.8 & 48.1 & 46.3 & 64.9 & 54.0\\
DreamLIP & 30M & 87.2 & 66.4 & 58.3 & 41.1 
&78.2 & 69.1 &66.5 & 47.4 & 59.3 & 56.1 & 49.1 & 47.9 &66.4 & 54.7\\

LLM2CLIP & 30M & 87.8 & 72.4 & 61.1 & 44.3 & 96.7 & \textbf{95.9} & 86.6 & 88.8 & \textbf{82.9} & 82.9 & 67.9 & 69.5 & 80.5 & 75.7 \\

\rowcolor[rgb]{0.925,0.957,1}\modelnm\ & 30M & \textbf{90.2} & \textbf{74.6} & \textbf{62.4} & \textbf{45.9} & \textbf{96.8} & \textbf{95.9} & \textbf{88.5} & \textbf{89.9} & \textbf{82.9} & \textbf{84.1} & \textbf{70.6} & \textbf{71.9} & \textbf{81.9(\textcolor{kcgreen}{+1.4})} & \textbf{77.0(\textcolor{kcgreen}{+1.3})} \\

\midrule
\multicolumn{6}{l}{\textit{Model Architecture: CLIP ViT-B/16}} & & & &  & & & & & \\

Long-CLIP & 1M & 90.0 & 76.2 & 62.8 & 46.3 & 88.4& 89.6 & 78.9&79.5 & 63.1 & 71.4 & 51.7 & 57.3 & 72.5 & 70.1  \\
FG-CLIP & 1.6B & 90.7 & 76.4 & 64.1 & 45.4 & 89.5 & 90.0 & 79.0 & 80.3 & 65.9 & 75.4 & 61.8 & 60.6 & 75.2 & 71.4 \\
\hdashline
LLM2CLIP & 3M & 88.0 & 75.3 & 60.5 & 44.8 & \textbf{94.4} & \textbf{94.4} & 80.6 & 86.0 & \textbf{81.7} & 82.2 & 67.2 & 69.1 & 78.7 & 75.3 \\
\rowcolor[rgb]{0.925,0.957,1}\modelnm\ & 3M & \textbf{89.4} & \textbf{77.6} & \textbf{61.7} & \textbf{46.8} & 94.3 & 93.3 & \textbf{82.9} & \textbf{88.1} & 81.0& \textbf{82.5} & \textbf{67.3} & \textbf{72.0} & \textbf{79.4(\textcolor{kcgreen}{+0.7})} & \textbf{76.7(\textcolor{kcgreen}{+1.4})} \\
\hdashline

LLM2CLIP & 15M & 88.9 & 76.6 & 62.4 & 46.5 & \textbf{95.0} & \textbf{95.2} & 84.5 & 88.4 & \textbf{83.8} & \textbf{85.1} & 69.3 & 72.4 & 80.7 & 77.3 \\
\rowcolor[rgb]{0.925,0.957,1}\modelnm\ & 15M & \textbf{90.8} & \textbf{77.9} & \textbf{63.2} &\textbf{47.8} & 94.2 & 94.9 &  \textbf{85.8} &\textbf{89.6} & 82.5 & 84.6 & \textbf{70.2} & \textbf{74.0} & \textbf{81.2(\textcolor{kcgreen}{+0.5})} & \textbf{78.0(\textcolor{kcgreen}{+0.7})} \\
\hdashline

CLIP & 30M & 90.5 & 75.9 & 63.2 & 48.2 
&77.9 & 68.3 &63.9 & 45.4 & 59.3 & 55.4 & 49.6 & 48.7 & 67.4 & 57.0\\
DreamLIP & 30M & 89.9 & 73.3 & 62.3 & 44.9 
&79.9 & 70.3 &67.9 & 50.1 & 60.4 & 57.3 & 50.2 & 48.8 &68.4 & 57.5\\

LLM2CLIP & 30M & 90.2 & 78.1 & 65.4 & 48.5 & \textbf{96.8} & \textbf{96.4} & 89.7 & 91.3 & \textbf{86.2} & 86.8 & 73.1 & 74.8 & 83.6 & 79.3 \\
\rowcolor[rgb]{0.925,0.957,1}\modelnm\ & 30M &\textbf{92.7} & \textbf{79.1} & \textbf{67.1} & \textbf{49.7} & 96.0 & \textbf{96.4} & \textbf{90.0} & \textbf{93.4} & 85.1 & \textbf{87.3} & \textbf{73.6} & \textbf{76.9} & \textbf{84.2(\textcolor{kcgreen}{+0.6})} & \textbf{80.5(\textcolor{kcgreen}{+1.2})} \\
\midrule
\multicolumn{6}{l}{\textit{Model Architecture: CLIP ViT-L/14}} & & & &  & & & & & \\
Long-CLIP & 1M &89.7 & 75.7 & 62.5 & 46.4 & 93.3 & 92.1 & 82.4 & 86.2 &  66.4 & 78.6 & 57.2 & 64.0 & 75.2 & 73.8     \\
FG-CLIP & 1.6B & 93.7 & 81.5 & 68.9 & 50.9 & 94.0  & 93.2 & 83.0 &  87.1 & 69.3 & 80.4 & 66.7 & 66.1 & 79.3 & 76.5\\
\hdashline
LLM2CLIP & 3M & 
92.4 & 80.1 & %
65.5 & 49.7 & %
\textbf{95.2} & \textbf{95.6} & %
83.6 & 89.0 & %
85.1 & 85.9 & %
70.0 & 74.4 & %
82.0  &79.1 \\ %

\rowcolor[rgb]{0.925,0.957,1}\modelnm\ & 3M & 
\textbf{92.8} & \textbf{81.1} & %
\textbf{66.4} & \textbf{51.9} & %
95.1 & 94.8 & %
\textbf{86.9} & \textbf{92.3} & %
\textbf{85.9} & \textbf{86.9} & %
\textbf{71.2} & \textbf{76.1} & %
 \textbf{83.0(\textcolor{kcgreen}{+1.0})} &\textbf{80.5(\textcolor{kcgreen}{+1.4})}  \\ %
 \hdashline

LLM2CLIP & 15M &  
91.3 & 80.6 & %
67.0& 50.6 & %
\textbf{96.3} &95.3 & %
86.4 & 90.5 & %
\textbf{86.4} & \textbf{88.5} & %
71.7 &75.3 & %
83.2 & 80.1 \\ %

\rowcolor[rgb]{0.925,0.957,1}\modelnm\ & 15M &  
\textbf{93.4} & \textbf{81.4} & %
\textbf{67.6}& \textbf{52.5} & %
96.1 & \textbf{95.4} & %
\textbf{88.3} & \textbf{92.6} & %
86.2 & 88.4 & %
\textbf{74.4} &\textbf{76.8} & %
\textbf{84.3(\textcolor{kcgreen}{+1.1})} & \textbf{81.2(\textcolor{kcgreen}{+1.1})} \\ %
\hdashline

LLM2CLIP & 30M & 
93.1 & 81.0 & %
68.2 & 52.0 & %
\textbf{97.5} & \textbf{97.7} & %
92.7 & 93.9 & %
\textbf{88.2} & 89.6 & %
74.9 & 78.3 & %
85.8 & 82.1\\ %

\rowcolor[rgb]{0.925,0.957,1}\modelnm\ & 30M & 
\textbf{94.5} & \textbf{81.6} & %
\textbf{69.3} & \textbf{53.2} & %
96.8 & 97.0 & %
\textbf{93.0} & \textbf{94.4} & %
87.5& \textbf{89.8} & %
\textbf{75.9} &\textbf{79.5} & %
\textbf{86.2(\textcolor{kcgreen}{+0.4})} & \textbf{82.6(\textcolor{kcgreen}{+0.5})} 
\\ %

\midrule
\multicolumn{6}{l}{\textit{MLLM-based method}} &  & & &  & & & & & \\
VLM2Vec (LLaVA Next 1.6) & 1M  &  91.6 & 79.8 & 61.3 & 51.5 & 93.9 & 91.0 & 90.0 & 92.0 & 80.5 & 86.0 & 68.7 & 74.7 & 81.0 & 79.2 \\
\bottomrule
\end{tabular}
}

\label{tab:retrieval}
\end{table*}

%% file: Table/zero_shot.tex
\begin{table*}[t!]
\centering
\caption{Zero-shot classification performance on 11 datasets. Bold numbers highlight the emphasized comparisons within each architecture block.}
\resizebox{\textwidth}{!}{
\begin{tabular}{lccccccccccccc}
\toprule
\textbf{Method}  &\textbf{\#Data} &  \rotatebox{90} {\textbf{Food101}} & \rotatebox{90} {\textbf{CIFAR10}} &  \rotatebox{90} {\textbf{CIFAR100}}& \rotatebox{90} {\textbf{SUN397}} & \rotatebox{90} {\textbf{Cars}} & \rotatebox{90} {\textbf{Aircraft}}& \rotatebox{90} {\textbf{DTD}} & \rotatebox{90} {\textbf{Pets}}  & \rotatebox{90} {\textbf{Caltech101}} &  \rotatebox{90} {\textbf{Flowers}}  & \rotatebox{90} {\textbf{ImageNet}} & \textbf{Avg.} \\
\midrule
\multicolumn{6}{l}{\textit{Model Architecture: CLIP ViT-B/32}} & & & &  & & & & \\

LLM2CLIP & 3M & 49.6 & 89.2 & 61.5 & 60.3 & 11.5 & 8.6 & 47.8 & 38.0 & 79.0 & 22.6 & 41.0 & 46.3\\
\rowcolor[rgb]{0.925,0.957,1}\modelnm\ & 3M & \textbf{64.5} & \textbf{90.7} & \textbf{65.8} & \textbf{65.0} & \textbf{21.2} & \textbf{11.6} & \textbf{52.0} & \textbf{51.7} & \textbf{83.3} & \textbf{30.8}  & \textbf{47.9} & \textbf{53.1(\textcolor{kcgreen}{+6.8})}\\
\hdashline
LLM2CLIP & 15M & 57.2 & 88.3 & 61.4 & 61.3 & 19.6 & 8.4 & 50.6 & 42.3 & 80.7 & 23.5 & 43.3 & 48.8 \\
\rowcolor[rgb]{0.925,0.957,1}\modelnm\ & 15M & \textbf{74.9} & \textbf{90.0} & \textbf{66.5} & \textbf{65.1} & \textbf{39.6} & \textbf{13.9} & \textbf{53.7} & \textbf{68.5} & \textbf{86.7} & \textbf{35.5} & \textbf{53.3} & \textbf{58.9(\textcolor{kcgreen}{+10.1})} \\
\hdashline
CLIP &30M&  54.8 & 90.0 & 67.1 & 62.0 & 13.0 & 3.6 & 27.6 & 49.4 & 83.8 & 41.4 & 45.6 & 48.9 \\

DreamLIP &30M&  68.2 & \textbf{91.8} & \textbf{69.2 }& 62.2 & 20.7 & 8.0 & 32.1 & 62.8 & 86.1 & \textbf{48.5} & \textbf{55.7} & 55.0 \\
LLM2CLIP & 30M & 58.5 & 88.3 & 61.0 & 61.2 & 20.6 & 8.4 & 50.3 & 37.6 & 81.7 & 26.0 & 45.1 & 49.0  \\
\rowcolor[rgb]{0.925,0.957,1}\modelnm\ & 30M & \textbf{74.4} & 88.8 & 66.9 & \textbf{65.9} & \textbf{38.0} & \textbf{16.2} & \textbf{53.0} & \textbf{64.5} & \textbf{86.8} & 40.4 & 54.0 & \textbf{59.0(\textcolor{kcgreen}{+10.0})} \\

\midrule
\multicolumn{6}{l}{\textit{Model Architecture: CLIP ViT-B/16}} & & & &  & & & & \\

Long-CLIP & 1M& 86.8 & 88.7 & 65.3 & 62.3 & 62.8 & 18.8 & 43.2 & 86.7 & 88.9 & 65.4 & 66.9 & 66.9 \\
FG-CLIP & 1.6B & 84.9 & 87.5 & 64.3 & 59.8  & 60.9& 17.6 & 40.3 & 85.7 & 86.6 & 64.4 & 65.8 & 65.3\\

\hdashline
LLM2CLIP & 3M & 56.9 & \textbf{92.6} & 64.4 & 62.2 & 15.4 & 11.7 & 50.9 & 46.5 & 82.9 & 23.6 & 45.8 & 50.3   \\

\rowcolor[rgb]{0.925,0.957,1}\modelnm\ & 3M & \textbf{73.1} & 92.5 & \textbf{68.9} & \textbf{67.9} & \textbf{32.3} & \textbf{13.5} & \textbf{54.1} & \textbf{59.8} & \textbf{87.0} & \textbf{35.8} & \textbf{54.8} & \textbf{58.2(\textcolor{kcgreen}{+7.9})}\\
\hdashline
LLM2CLIP & 15M & 63.2 & \textbf{90.8} & 64.5 & 62.9 & 27.3 & 9.9 & 52.8 & 50.3 & 83.2 & 23.7 & 46.5 & 52.3   \\
\rowcolor[rgb]{0.925,0.957,1}\modelnm\ & 15M & \textbf{80.3} & \textbf{90.8} & \textbf{69.7} & \textbf{67.4} & \textbf{44.3} & \textbf{16.5} & \textbf{56.7} & \textbf{75.8} & \textbf{88.4} & \textbf{40.8} & \textbf{58.6} & \textbf{62.7(\textcolor{kcgreen}{+10.4})}\\
\hdashline
 CLIP & 30M & 61.3 & 92.2 & 66.9 & 62.2 & 19.3 & 5.7 & 30.9 & 49.3 & 83.7 & 43.4 & 50.0 & 51.4 \\

 DreamLIP & 30M & 76.4 & \textbf{92.7} & \textbf{71.7 }& 64.7 & 23.4 & 8.3 & 33.9 & 66.3 & 87.6 &\textbf{59.1} & 58.4 & 58.4 \\
LLM2CLIP & 30M & 64.4 & 90.2 & 64.6 & 63.7 & 27.0 & 11.2 & 55.0 & 45.9 & 84.0 & 27.1 & 49.7 & 53.0\\
\rowcolor[rgb]{0.925,0.957,1}\modelnm\ & 30M & \textbf{81.0} & 89.3 & 68.3 & \textbf{68.2} & \textbf{48.5} & \textbf{17.9} & \textbf{57.3} & \textbf{70.2} & \textbf{88.8} & 44.8 & \textbf{59.2} & \textbf{63.0(\textcolor{kcgreen}{+10.0})} \\

\midrule
\multicolumn{6}{l}{\textit{Model Architecture: CLIP ViT-L/14}} & & & &  & & & & \\

Long-CLIP & 1M & 90.8 & 96.3 & 78.7 & 71.3 & 70.3 & 28.1 & 56.0 & 90.4 & 93.6 & 71.1 & 72.0 & 74.5\\
FG-CLIP & 1.6B & 89.7 & 95.7 & 76.9 & 69.9 & 68.4 & 26.8 & 55.7 & 79.8 & 92.1 & 69.8 & 70.4 & 72.3\\
\hdashline
LLM2CLIP & 3M & 64.8 & 95.4 & 72.9 & 66.4 & 18.8 & 10.4 & 54.8 & 47.3 & 88.3 & 26.8 & 52.8  & 54.4 \\
\rowcolor[rgb]{0.925,0.957,1}\modelnm\ & 3M & \textbf{83.4} & \textbf{96.6} & \textbf{78.3} & \textbf{72.4} & \textbf{45.1} & \textbf{16.2} & \textbf{59.6} & \textbf{65.9} & \textbf{92.3} & \textbf{41.8} & \textbf{62.5} & \textbf{64.9 (\textcolor{kcgreen}{+10.5})} \\
\hdashline
LLM2CLIP & 15M &  70.1 & 95.2 & 72.3 & 66.4 & 32.4 & 9.5 & 58.0 & 54.3 & 88.3 & 26.6 & 54.0 &57.0 \\
\rowcolor[rgb]{0.925,0.957,1}\modelnm\ & 15M & \textbf{87.1} & \textbf{95.4} & \textbf{77.6} & \textbf{72.3} & \textbf{59.8} & \textbf{21.1} & \textbf{62.1} & \textbf{77.0} & \textbf{92.4} & \textbf{48.8} & \textbf{66.0} & \textbf{69.3 (\textcolor{kcgreen}{+12.3})} \\
 \hdashline
LLM2CLIP & 30M & 71.2 & 94.0 & 70.5 & 67.0 & 32.1 & 11.3 & 57.8 & 54.7 & 89.3 & 28.8 & 56.4 & 57.5\\
\rowcolor[rgb]{0.925,0.957,1}\modelnm\ & 30M & \textbf{88.9} &\textbf{ 94.1} & \textbf{77.7} & \textbf{72.5} & \textbf{61.1} & \textbf{25.2} & \textbf{62.8} & \textbf{81.5} & \textbf{92.9} & \textbf{57.2} & \textbf{67.8} & \textbf{71.0 (\textcolor{kcgreen}{+13.5})} \\

\bottomrule
\end{tabular}
}
\vspace{-2mm}
\label{tab:classification}
\end{table*}

%% file: Table/llava.tex
\begin{table*}[t!]
    \centering
        \caption{MLLM performance under $224^2$ image resolution.}
    \resizebox{1.0\textwidth}{!}{
    \begin{tabular}{lccccccc}
    \toprule
        \textbf{Method} & \textbf{SEED-Bench (image)} & \textbf{GQA} & \textbf{VizWiz} & \textbf{PoPE} & \textbf{TextVQA} & \textbf{MMBench} & \textbf{VQAv2} \\
        \midrule
CLIP & \textcolor{gray}{65.3} & \textcolor{gray}{62.0} & \textcolor{gray}{44.0} &
\textcolor{gray}{85.7} & \textcolor{gray}{54.2} & \textcolor{gray}{65.5} & \textcolor{gray}{77.4}\\
    \rowcolor[rgb]{0.925,0.957,1}
        ProCLIP (ViT-L-30M)  & \textbf{66.4} & \textbf{62.3} & \textbf{44.7} & \textbf{85.9} & \textbf{54.5
    }& 65.0& \textbf{78.1}\\
        \bottomrule
    \end{tabular}
}

    \vspace{-3mm}
    \label{tab:llava}
\end{table*}

%% file: Section/conclusion.tex
\section{Conclusion}
\vspace{-2mm}
This paper presents \textbf{\modelnm}, a progressive vision-language alignment framework that integrates CLIP image encoders with LLM-based embedders. \modelnm\ employs a two-stage strategy: first, anchoring the LLM embedder to CLIP's textual space via knowledge distillation to inherit semantic priors; second, performing cross-modal contrastive tuning with self-distillation to prevent catastrophic forgetting. To maintain feature-space consistency, we introduce instance-level semantic and global structural alignment constraints. Extensive evaluations across various scales and architectures validate the efficacy and generality of \textbf{\modelnm}.  We will release all model weights and code.

%% file: checklist.tex
\section*{NeurIPS Paper Checklist}

\begin{enumerate}

\item {\bf Claims}
    \item[] Question: Do the main claims made in the abstract and introduction accurately reflect the paper's contributions and scope?
    \item[] Answer: \answerYes{} %
    \item[] Justification: The abstract and introduction clearly explain the research background and motivation of ProCLIP, as well as the overall rationale behind its methodological design. They also show that ProCLIP significantly outperforms the baseline across multiple aspects and establishes a new state-of-the-art in the field of LLM-enhanced CLIP.

\item {\bf Limitations}
    \item[] Question: Does the paper discuss the limitations of the work performed by the authors?
    \item[] Answer: \answerYes{} %
    \item[] Justification: In Appendix F, Future Work, we clearly discuss the current limitations of ProCLIP, including its increased computational overhead. We also point out several possible directions for addressing this issue.

\item {\bf Theory assumptions and proofs}
    \item[] Question: For each theoretical result, does the paper provide the full set of assumptions and a complete (and correct) proof?
    \item[] Answer: \answerYes{} %
    \item[] Justification: In Appendix C, Theoretical Analysis, we provide a thorough theoretical discussion of the design of ProCLIP, along with complete assumptions and proofs.

    \item {\bf Experimental result reproducibility}
    \item[] Question: Does the paper fully disclose all the information needed to reproduce the main experimental results of the paper to the extent that it affects the main claims and/or conclusions of the paper (regardless of whether the code and data are provided or not)?
    \item[] Answer: \answerYes{} %
    \item[] Justification: In Appendix B, Experiment Details, we present all hyperparameters used in this work. We also clearly describe the datasets and base models used in the main paper.

\item {\bf Open access to data and code}
    \item[] Question: Does the paper provide open access to the data and code, with sufficient instructions to faithfully reproduce the main experimental results, as described in supplemental material?
    \item[] Answer: \answerNo{} %
    \item[] Justification: Our codebase is entirely built upon the publicly available LLM2CLIP codebase. We will also fully open-source our own code in the near future, including the model weights.

\item {\bf Experimental setting/details}
    \item[] Question: Does the paper specify all the training and test details (e.g., data splits, hyperparameters, how they were chosen, type of optimizer) necessary to understand the results?
    \item[] Answer: \answerYes{} %
    \item[] Justification: In Appendix B, Experiment Details, we clearly describe the training and evaluation details. The datasets used in this work are also clearly described in the main paper.

\item {\bf Experiment statistical significance}
    \item[] Question: Does the paper report error bars suitably and correctly defined or other appropriate information about the statistical significance of the experiments?
    \item[] Answer: \answerNo{} %
    \item[] Justification: We did not report a statistical significance analysis due to the high computational cost of the experiments. However, it is worth noting that the training datasets, training code, and random seeds are fixed, which helps ensure the reproducibility of our experiments.

\item {\bf Experiments compute resources}
    \item[] Question: For each experiment, does the paper provide sufficient information on the computer resources (type of compute workers, memory, time of execution) needed to reproduce the experiments?
    \item[] Answer: \answerYes{} %
    \item[] Justification: In Appendix B, Experiment Details, we clearly state that the experiments were conducted using 8 $\times$H100 GPUs. 
\item {\bf Code of ethics}
    \item[] Question: Does the research conducted in the paper conform, in every respect, with the NeurIPS Code of Ethics \url{https://neurips.cc/public/EthicsGuidelines}?
    \item[] Answer: \answerYes{} %
    \item[] Justification: The authors have reviewed the NeurIPS Code of Ethics and confirm that the research conforms to it. The paper preserves submission anonymity, uses existing research assets with proper citation, and does not involve human subjects or private sensitive data

\item {\bf Broader impacts}
    \item[] Question: Does the paper discuss both potential positive societal impacts and negative societal impacts of the work performed?
    \item[] Answer: \answerNo{} %
    \item[] Justification: The topic studied in this paper is purely within the field of machine learning and does not have any obvious additional positive or negative societal impact. Its primary value lies in the machine learning community, as this work provides a superior algorithm.
    
\item {\bf Safeguards}
    \item[] Question: Does the paper describe safeguards that have been put in place for responsible release of data or models that have a high risk for misuse (e.g., pre-trained language models, image generators, or scraped datasets)?
    \item[] Answer: \answerNA{} %
    \item[] Justification:  This work focuses on CLIP-style vision-language representation learning rather than releasing high-risk generative models or scraped datasets. The released code and model weights are intended for research purposes, so this question is not directly applicable.

\item {\bf Licenses for existing assets}
    \item[] Question: Are the creators or original owners of assets (e.g., code, data, models), used in the paper, properly credited and are the license and terms of use explicitly mentioned and properly respected?
    \item[] Answer: \answerYes{} %
    \item[] Justification:  We properly credit the original creators of the codebase, datasets, and base models used in this paper. We also follow their corresponding licenses and terms of use.

\item {\bf New assets}
    \item[] Question: Are new assets introduced in the paper well documented and is the documentation provided alongside the assets?
    \item[] Answer: \answerYes{} %
    \item[] Justification:  The newly introduced assets, including code and model weights, will be released with accompanying documentation, usage instructions, and evaluation details.

\item {\bf Crowdsourcing and research with human subjects}
    \item[] Question: For crowdsourcing experiments and research with human subjects, does the paper include the full text of instructions given to participants and screenshots, if applicable, as well as details about compensation (if any)? 
    \item[] Answer: \answerNA{} %
    \item[] Justification: This paper does not involve crowdsourcing experiments or research with human subjects, so participant instructions, screenshots, and compensation details are not applicable.

\item {\bf Institutional review board (IRB) approvals or equivalent for research with human subjects}
    \item[] Question: Does the paper describe potential risks incurred by study participants, whether such risks were disclosed to the subjects, and whether Institutional Review Board (IRB) approvals (or an equivalent approval/review based on the requirements of your country or institution) were obtained?
    \item[] Answer: \answerNA{} %
    \item[] Justification: This paper does not involve human subjects or study participants, so potential participant risks and IRB approval are not applicable.

\item {\bf Declaration of LLM usage}
    \item[] Question: Does the paper describe the usage of LLMs if it is an important, original, or non-standard component of the core methods in this research? Note that if the LLM is used only for writing, editing, or formatting purposes and does \emph{not} impact the core methodology, scientific rigor, or originality of the research, declaration is not required.
    \item[] Answer: \answerYes{} %
    \item[] Justification: The paper describes how LLMs are used as a core component of ProCLIP to enhance CLIP-style vision-language representation learning.

\end{enumerate}

%% file: Section/appendix.tex
\addcontentsline{toc}{section}{Appendix}

\noindent

The appendix is organized as follows:

\begin{itemize}

    \item \textbf{Appendix A: Methodological Discussion.} 

    We discuss the differences between ProCLIP and representative CLIP extension methods, including Long-CLIP, FG-CLIP, and LLM2CLIP.

    \item \textbf{Appendix B: Experimental Details.} 

    We provide detailed training settings, evaluation protocols, benchmark descriptions, dataset statistics, and prompt templates.

    \item \textbf{Appendix C: Theoretical Analysis.} 

    We theoretically analyze why progressive alignment can incorporate LLM-based textual representations into CLIP while preserving its pre-trained generalization ability.

    \item \textbf{Appendix D: More Results and Analysis.} 

    We report additional quantitative results, including fine-grained understanding, multilingual retrieval, linear probing, robustness, data/model scaling, open-vocabulary semantic segmentation, and efficiency analysis.

    \item \textbf{Appendix E: Visualization.} 

    We present qualitative visualizations to demonstrate the advantages of ProCLIP in long-text cross-modal retrieval.

    \item \textbf{Appendix F: Future Work.} 

    We discuss several potential directions for improving ProCLIP, including training efficiency, fine-grained alignment, and architecture extension.

\end{itemize}

\section{Methodological Discussion}
Although several recent methods have attempted to enhance CLIP from different perspectives, ProCLIP differs from them in both motivation and capability coverage. Long-CLIP mainly focuses on extending the text length supported by CLIP, while FG-CLIP emphasizes fine-grained visual-textual alignment and recognition. These methods are effective in preserving or even improving zero-shot classification performance, but their improvements are relatively specialized. In contrast, ProCLIP aims to provide a more comprehensive extension of CLIP by aligning it with stronger LLM-based textual representations. Beyond maintaining competitive recognition ability, ProCLIP further improves cross-modal retrieval, multilingual understanding, long-text comprehension, and fine-grained semantic alignment.

Another important distinction lies in training efficiency. Compared with LLM2CLIP, which directly aligns CLIP with LLM representations and may severely damage the original visual generalization ability, ProCLIP introduces a curriculum alignment strategy with self-distillation regularization. This design enables stable cross-modal alignment while preserving the pre-trained knowledge of CLIP. As a result, ProCLIP achieves a better trade-off between enhancing advanced language-side capabilities and retaining the generalization ability of the original vision-language model. Therefore, ProCLIP should not be viewed as a method solely optimized for zero-shot classification, but rather as a general-purpose CLIP enhancement framework for diverse cross-modal scenarios.
\begin{table}[h]
\centering
\caption{Comparison between ProCLIP and representative CLIP extension methods. ProCLIP provides a more comprehensive enhancement across multilingual, long-text, fine-grained scenarios while maintaining data-efficient training.}
\label{tab:method_comparison}
\resizebox{\linewidth}{!}{
\begin{tabular}{lcccccc}
\toprule
\textbf{Method} & 
\textbf{Training Data} & 
\textbf{Training Method} & 
\textbf{Long Text} & 
\textbf{Multilingual} & 
\textbf{Fine-grained} & 
\textbf{Zero-shot Recognition} \\
\midrule
CLIP & Large-scale & Pre-train & Limited & Limited & Moderate  & Strong \\
Long-CLIP & Small-scale &Post-train & \checkmark & Limited  & Moderate & Strong \\
FG-CLIP & Large-scale &Continue Pre-train & Limited & Limited & \checkmark  & Strong \\
LLM2CLIP & Small-scale  &Post-train& \checkmark & \checkmark & \checkmark  & Moderate \\
ProCLIP & Small-scale  &Post-train & \checkmark & \checkmark & \checkmark &  Strong \\
\bottomrule
\end{tabular}
}
\end{table}

\section{Experiment Details}
\subsection{Training Details} 
Details of the hyperparameter configurations used for two-stage training of ProCLIP are presented in Tab.~\ref{tab:parma}. Under the default setting, our MLP layers are consistent with the baseline LLM2CLIP, both consisting of four linear layers. 

\begin{table}[h!]
    \centering
    \caption{Detailed hyperparameters for training ProCLIP.}
    \captionsetup{font=small}
    \begin{minipage}{0.5\linewidth}
        \centering
        \captionsetup{font=small}
        \resizebox{\linewidth}{!}{
        \begin{tabular}{ll}
        \toprule
        \multicolumn{2}{c}{\textbf{Hyperparameters of stage1}} \\
        \midrule
        Batch size & 1024 ($8\times128$) \\
        Optimizer & AdamW\\
        Weight decay & 0.05\\
        Adam $\beta$ & (0.9,0.98)\\
        Adam $\epsilon$ & 1e-6\\
        Learning rate & 1e-5 \\
        Learning rate schedule & cosine decay\\
        Epochs & 4 \\
        Training GPUs & $8\times$H100\\
        \bottomrule
        \end{tabular}}
    \end{minipage}
    \hfill
    \begin{minipage}{0.43\linewidth}
        \centering
        \captionsetup{font=small}
        \resizebox{\linewidth}{!}{
        \begin{tabular}{ll}
        \toprule
        \multicolumn{2}{c}{\textbf{Hyperparameters of stage2}} \\
        \midrule
        Batch size & 4096 ($8\times512$) \\
        Optimizer & AdamW\\
        Weight decay & 0.05\\
        Adam $\beta$ & (0.9,0.98)\\
        Adam $\epsilon$ & 1e-6\\
        Learning rate & 1e-5 \\
        Learning rate schedule & cosine decay\\
        Ema $\alpha$ & 0.999\\
        $\lambda_{\text{reg}}$ &0.0004\\
        Epochs & 4 \\
        Training GPUs & $8\times$H100\\
        \bottomrule
        \end{tabular}}
    \end{minipage}

    \label{tab:parma}
\end{table}
\subsection{Evaluation Details} 
For the LLM2CLIP baseline, we reproduce all results with the official LLM2CLIP codebase, since LLM2CLIP does not provide a complete set of checkpoints from ViT-B/32 to ViT-L/14 and released checkpoints may be affected by different hardware and training environments. For a fair comparison, all LLM2CLIP results reported in the main paper are reproduced by ourselves using the same training data as ProCLIP and training for 4 epochs. We will release all checkpoints in the future, including both our reproduced baselines and ProCLIP.

For the ShareGPT4V benchmark, since no fixed evaluation set is provided, we randomly sample 1,000 images from COCO2017, following the common practice in prior works.

For comparison methods, we mainly compare ProCLIP with representative works, including CLIP~\cite{clip}, LLM2CLIP~\cite{llm2clip}, DreamLIP~\cite{dreamlip}, FG-CLIP~\cite{fgclip}, and Long-CLIP~\cite{longclip}. Among them, LLM2CLIP serves as our primary baseline. Notably, LLM2CLIP received the Outstanding Paper Award at AAAI 2026. Building upon it, we propose ProCLIP, a more comprehensive and effective method.
\subsection{Details of Benchmarks.}
\noindent\textbf{Zero-Shot Classification \& Linear Probe.} 
Following the previous works~\cite{alip,rwkvclip}, we evaluate the zero-shot classification and linear probe performance of the models on 11 datasets. The detailed information about these datasets and the prompt used in zero-shot classification are presented in Tab.~\ref{tab:linearprobe_datasets} and Tab.~\ref{tab:prompt}.

\begin{table}[h!]
\centering
\caption{List of zero-shot datasets with the data distribution and evaluation metrics.}
\captionsetup{font=small}

\begin{tabular}{lcccr}
\toprule
\multicolumn{1}{l}{Dataset} & \multicolumn{1}{c}{Classes} & \multicolumn{1}{c}{Train size} & \multicolumn{1}{c}{Test size} & \multicolumn{1}{c}{Evaluation metric} \\
\midrule
Food101                        & 101                             & 75,750                             & 25,250                            & accuracy                                  \\
CIFAR10                        & 10                              & 50,000                             & 10,000                            & accuracy                                  \\
CIFAR100                       & 100                             & 50,000                             & 10,000                            & accuracy                                  \\
SUN397                          & 397                             & 19,850                             & 19,850                            & accuracy                                  \\
Cars                   & 196                             & 8,144                              & 8,041                             & accuracy                                  \\
Aircraft                   & 100                             & 6,667                              & 3,333                             & mean per class                            \\
DTD           & 47                              & 3,760                              & 1,880                             & accuracy                                  \\
Pets                & 37                              & 3,680                              & 3,669                             & mean per class                            \\
Caltech101                     & 101                             & 3,000                              & 5,677                             & mean-per-class                            \\
Flowers                  & 102                             & 2,040                              & 6,149                             & mean per class                            \\

ImageNet                        & 1000                            & 1,281,167                          & 50,000                            & accuracy                                  \\
\bottomrule
\end{tabular}
\label{tab:linearprobe_datasets}
\end{table}

\noindent\textbf{Robustness.} We evaluated the robustness of our model on five out-of-distribution datasets, including ImageNet-v2~\cite{IN_v2}, ImageNet-A~\cite{IN_a}, ImageNet-O~\cite{IN_a}, ImageNet-R~\cite{IN_r}, and ImageNet-Sketch~\cite{IN_s}.

\noindent\textbf{Cross-Modal Retrieval.} 
Following the previous works~\cite{llm2clip,flame}, we evaluate the cross-modal retrieval performance of the models on 6 datasets: Flickr30k~\cite{flickr30k}, COCO~\cite{coco}, ShareGPT4V~\cite{sharegpt4v}, Urban-1k~\cite{longclip}, DOCCI~\cite{docci}, and DCI~\cite{dci}. The details information about these dataset are present on Tab.~\ref{tab:retrieval_settings}.

\begin{table}[h!]
\centering
\caption{Zero-shot image-text retrieval evaluation settings.}
\captionsetup{font=small}

\begin{tabular}{lccc}
\toprule
\textbf{Dataset} & \textbf{Test Images} & \textbf{Evaluation Protocol} & \textbf{Text type} \\
\midrule
MSCOCO & 5,000  & Image-to-Text \& Text-to-Image & short \\
Flickr30k & 1,000  & Image-to-Text \& Text-to-Image & short \\
ShareGPT4V & 1000  & Image-to-Text \& Text-to-Image & long \\
Urban-1k & 1000 & Image-to-Text \& Text-to-Image & long \\
DOCCI & 5000 &  Image-to-Text \& Text-to-Image & long \\
DCI & 7805 & Image-to-Text \& Text-to-Image & long \\
\bottomrule
\end{tabular}

\vspace{-3mm}
\label{tab:retrieval_settings}
\end{table}
\input{Table/sugar_crepe_tongji}

\noindent\textbf{Multilingual Retrieval.} We evaluated the multilingual capabilities of our model on XM3600~\cite{thapliyal2022crossmodal}. XM3600 contains 3,600 images covering a total of 36 languages, including Arabic (ar), Bengali(bn), Chinese-Simplified (zh), Croatian (hr), Czech (cs), Danish (da), Dutch (nl), English (en),Farsi (fa), Filipino (fil), Finnish (fi), French (fr), German (de), Greek (el), Hebrew (he), Hindi (hi), Hungarian (hu), Indonesian (id), Italian (it), Japanese (ja), Korean (ko),Maori(mi), Norwegian (no), Persian (fa), Polish (pl), Portuguese (pt), Romanian (ro), Russian (ru), Spanish (es), Swedish (sv), Swahili(sw), Thai (th), Turkish (tr), Telugu (te), Ukrainian (uk), and Vietnamese (vi).
\begin{table}[t!]
\centering
\caption{Full list of prompts to evaluate the performance of zero-shot classification on 11 visual recognition datasets.}
\captionsetup{font=small}
\resizebox{1.0\linewidth}{!}{
\begin{tabular}{llll}
\toprule
\multicolumn{4}{l}{\bf CIFAR 10 \& CIFAR 100} \\
a photo of a \{label\}. &
a blurry photo of a \{label\}. &
a black and white photo of a \{label\}. &
a low contrast photo of a \{label\}. \\
a high contrast photo of a \{label\}. &
a bad photo of a \{label\}. &
a good photo of a \{label\}. &
a photo of a small \{label\}. \\
a photo of a big \{label\}.&
a photo of the \{label\}.&
a blurry photo of the \{label\}.&
a black and white photo of the \{label\}. \\
a low contrast photo of the \{label\}.&
a high contrast photo of the \{label\}.&
a bad photo of the \{label\}.&
a good photo of the \{label\}. \\
a photo of the small \{label\}.&
a photo of the big \{label\}.& & \\
\midrule
\multicolumn{4}{l}{\bf Food101} \\
a photo of \{label\}, a type of food. & & \\
\midrule
\multicolumn{4}{l}{\bf Caltech101} \\
a photo of a \{label\}. &
a painting of a \{label\}. &
a plastic \{label\}. &
a sculpture of a \{label\}. \\
a sketch of a \{label\}. &
a tattoo of a \{label\}. &
a toy \{label\}. &
a rendition of a \{label\}. \\
a embroidered \{label\}. &
a cartoon \{label\}. &
a \{label\} in a video game. &
a plushie \{label\}. \\
an origami \{label\}. &
art of a \{label\}. &
graffiti of a \{label\}. &
a drawing of a \{label\}. \\
a doodle of a \{label\}. &
a photo of the \{label\}. &
a painting of the \{label\}.&
the plastic \{label\}. \\
a sculpture of the \{label\}.&
a sketch of the \{label\}.&
a tattoo of the \{label\}.&
the toy \{label\}. \\
a rendition of the \{label\}.&
the embroidered \{label\}.&
the cartoon \{label\}.&
the \{label\} in a video game. \\
the plushie \{label\}.&
the origami \{label\}.&
art of the \{label\}.&
graffiti of the \{label\}. \\
a drawing of the \{label\}.&
a doodle of the \{label\}.& & \\
\midrule
\multicolumn{4}{l}{\bf Stanford Cars} \\
a photo of a \{label\}.&
a photo of the \{label\}.&
a photo of my \{label\}.&
i love my \{label\}! \\
a photo of my dirty \{label\}.&
a photo of my clean \{label\}.&
a photo of my new \{label\}.&
a photo of my old \{label\}. \\
\midrule
\multicolumn{4}{l}{\bf DTD} \\
a photo of a \{label\} texture.&
a photo of a \{label\} pattern.&
a photo of a \{label\} thing.&
a photo of a \{label\} object. \\
a photo of the \{label\} texture. &
a photo of the \{label\} pattern. &
a photo of the \{label\} thing. &
a photo of the \{label\} object. \\
\midrule
\multicolumn{4}{l}{\bf FGVC Aircraft} \\
a photo of a \{label\}, a type of aircraft.&
a photo of the \{label\}, a type of aircraft.& & \\
\midrule
\multicolumn{4}{l}{\bf Flowers102} \\
a photo of a \{label\}, a type of flower. &&& \\
\midrule
\multicolumn{4}{l}{\bf Pets } \\
a photo of a \{label\}, a type of pet.&&& \\
\midrule
\multicolumn{4}{l}{\bf  SUN39} \\
a photo of a \{label\}.&
a photo of the \{label\}.&& \\
\midrule
\multicolumn{4}{l}{\bf  ImageNet} \\
a bad photo of a \{label\}. & 
a photo of many \{label\}. &
a sculpture of a \{label\}. &
a photo of the hard to see \{label\}. \\
a low resolution photo of the \{label\}. & 
a rendering of a \{label\}. &
graffiti of a \{label\}. &
a bad photo of the \{label\}.  \\
a cropped photo of the \{label\}. &
a tattoo of a \{label\}. & 
the embroidered \{label\}. &
a photo of a hard to see \{label\}.  \\
a bright photo of a \{label\}.&
a photo of a clean \{label\}.&
a photo of a dirty \{label\}.&
a dark photo of the \{label\}. \\
a drawing of a \{label\}.&
a photo of my \{label\}.&
the plastic \{label\}.&
a photo of the cool \{label\}. \\
a close-up photo of a \{label\}.&
a black and white photo of the \{label\}.&
a painting of the \{label\}.&
a painting of a \{label\}. \\
a pixelated photo of the \{label\}.& 
a sculpture of the \{label\}.&
a bright photo of the \{label\}.&
a cropped photo of a \{label\}. \\
a plastic \{label\}.&
a photo of the dirty \{label\}.& 
a jpeg corrupted photo of a \{label\}.&
a blurry photo of the \{label\}. \\
a photo of the \{label\}.&
a good photo of the \{label\}.&
a rendering of the \{label\}.&
a \{label\} in a video game. \\
a photo of one \{label\}.&
a doodle of a \{label\}.&
a close-up photo of the \{label\}.&
a photo of a \{label\}. \\
the origami \{label\}.&
the \{label\} in a video game.&
a sketch of a \{label\}.&
a doodle of the \{label\}. \\
an origami \{label\}.&
a low resolution photo of a \{label\}.&
the toy \{label\}.&
a rendition of the \{label\}. \\
a photo of the clean \{label\}.& 
a photo of a large \{label\}.& 
a rendition of a \{label\}.&
a photo of a nice \{label\}. \\
a photo of a weird \{label\}.& 
a blurry photo of a \{label\}.&
a cartoon \{label\}.&
art of a \{label\}. \\
a sketch of the \{label\}.& 
a embroidered \{label\}.&
a pixelated photo of a \{label\}.&
itap of the \{label\}. \\
a jpeg corrupted photo of the \{label\}.& 
a good photo of a \{label\}.&
a plushie \{label\}.&
a photo of the nice \{label\}. \\
a photo of the small \{label\}.& 
a photo of the weird \{label\}.&
the cartoon \{label\}.&
art of the \{label\}. \\
a drawing of the \{label\}.& 
a photo of the large \{label\}.& 
a black and white photo of a \{label\}.&
the plushie \{label\}. \\
a dark photo of a \{label\}.& 
itap of a \{label\}.& 
graffiti of the \{label\}.& 
a toy \{label\}. \\
itap of my \{label\}.& 
a photo of a cool \{label\}.&
a photo of a small \{label\}.& 
a tattoo of the \{label\}. \\
\bottomrule
\end{tabular}}
\label{tab:prompt}
\end{table}

\noindent\textbf{Fine-Grained Understanding.} 
We evaluated the fine-grained understanding capability of the VLM on MMVP-VLM~\cite{mmvp} and SugarCrepe. Here, fine-grained understanding is conceptually similar to compositional understanding, focusing primarily on the model’s ability to distinguish subtle textual differences in cross-modal matching. MMVP-VLM consists of 150 samples in total, testing 9 patterns:
\begin{itemize}
    \item \textbf{\faCompass \text{ }Orientation and Direction}: Questions about the direction something is facing or moving, such as the direction the dog or duck is facing, or the orientation of the school bus.
    \item \textbf{\faSearch \text{ }Presence of Specific Features}: Questions that focus on the existence or non-existence of certain elements or features in the image.
    \item \textbf{\faSync\text{ } State and Condition}: Questions that pertain to the state or condition of an object, such as whether a flag is blowing in the wind or if the ground is wet.
    \item \textbf{\faSortNumericUp \text{ }Quantity and Count}: Questions about the number of objects or features present in the image.
    \item \textbf{\faMapPin\text{ } Positional and Relational Context}: This aspect refers to the model's ability to understand the position and relationship of objects or elements within an image in relation to each other and their surroundings. 
    \item \textbf{\faPalette\text{ } Color and Appearance}: Questions regarding the color of certain objects or elements.
    \item \textbf{\faCogs\text{ } Structural and Physical Characteristics}: This category involves the model's ability to identify and analyze the physical attributes and structural features of objects in an image. 
    \item \textbf{\faFont\text{ } Text}: Questions related to text or symbols present in the image.
    \item \textbf{\faCamera \text{ }Viewpoint and Perspective}: Questions concerning the perspective from which the photo was taken.
\end{itemize}
SugarCrepe contains three types of negative samples—Add, Replace, and Swap—designed to probe whether vision-language models (VLMs) can recognize text that differs only in subtle structural changes yet conveys a completely different overall meaning in image-text matching tasks.

\noindent\textbf{MLLM benchmarks.}
We further integrate the fine-tuned vision encoder into LLaVA and evaluate its performance on several MLLM downstream benchmarks, including SEED-Bench~\cite{seed-bench}, GQA~\cite{gqa}, VizWiz~\cite{vizwiz}, PoPE~\cite{pope}, TextVQA~\cite{textvqa}, MMBench~\cite{mmbench}, and VQAv2~\cite{vqav2}.

\noindent\textbf{Open-Vocabulary Semantic Segmentation.} 
We further evaluate the model performance on a pixel-level task, namely open-vocabulary semantic segmentation, across three major datasets: ADE20K~\cite{ade20k}, COCO-Stuff164k~\cite{cocostuff}, and VOC20~\cite{voc}. We adopt the same evaluation protocol as MaskCLIP~\cite{maskclip}.

\section{Theoretical Analysis}
\label{sec:theoretical_analysis}

In this section, we provide a theoretical interpretation of why the proposed progressive alignment framework can effectively incorporate LLM representations into CLIP while preserving the pre-trained generalization ability of CLIP. Our analysis focuses on four key questions: 
(1) why directly aligning an uncalibrated LLM space with the visual space may cause representation degradation; 
(2) how instance-level and structural distillation jointly provide local and global semantic constraints across the two training stages; 
(3) why short-text distillation can benefit long-text understanding; 
and (4) why multilingual capability can emerge even without multilingual image-text pairs during training.

Let $\mathcal{V}$ denote the visual embedding space and $\mathcal{T}$ denote the CLIP text embedding space. For each training instance, let $v_i \in \mathcal{V}$ be the visual embedding, $e_i \in \mathcal{T}$ be the original CLIP text embedding, and $t_i^* \in \mathcal{T}$ be the projected LLM embedding produced by the adapter. We assume that all embeddings are normalized, as commonly adopted in CLIP-like contrastive learning.

\subsection{Why Direct Alignment May Degrade CLIP Representations}

Existing direct alignment methods, such as LLM2CLIP, typically optimize the contrastive objective between visual embeddings and projected LLM embeddings without first ensuring that the LLM space is compatible with the original CLIP text space. In this case, the training process needs to simultaneously discover a proper language-to-CLIP transformation and perform cross-modal alignment. This may introduce unstable gradients, especially when the projected LLM embeddings are initially far away from the CLIP semantic manifold.

Let $\mathcal{L}_{\text{info}}$ denote the InfoNCE loss. For a positive image-text pair $(v_i,t_i^*)$, the contrastive gradient encourages the visual embedding $v_i$ to move toward the projected LLM embedding $t_i^*$ while pushing it away from other text embeddings. However, when $t_i^*$ has not yet inherited the geometry of the original CLIP text space, this update direction may deviate from the well-structured visual-textual alignment learned by CLIP. Consequently, the visual encoder may be forced to fit an uncalibrated or geometrically incompatible target space, which can damage its pre-trained zero-shot generalization ability.

Formally, suppose the original CLIP model has learned a well-aligned cross-modal structure satisfying
\begin{equation}
    \mathbb{E}\left[d(v_i^{(0)}, e_i)\right] \leq \epsilon_C,
\end{equation}
where $v_i^{(0)}$ denotes the visual embedding extracted by the original CLIP encoder and $d(\cdot,\cdot)$ is a metric distance. In direct alignment, the projected LLM embedding $t_i^*$ is optimized mainly through the image-text contrastive loss. Before the adapter becomes well calibrated, the discrepancy
\begin{equation}
    d(v_i^{(0)}, t_i^*)
\end{equation}
is not explicitly controlled. Therefore, the gradient of $\mathcal{L}_{\text{info}}$ may drive the visual representation away from its original CLIP neighborhood.

This explains why direct alignment may lead to representation degradation: the visual encoder is updated toward an LLM-induced target space that has not yet been aligned with the CLIP text manifold. As a result, the optimization may partially overwrite the original CLIP representation rather than merely refining it. This phenomenon is particularly harmful under limited training data, where the model can easily overfit to the alignment data and lose the broad zero-shot transfer ability acquired during CLIP pre-training.

\subsection{Dual Distillation as Inheritance and Preservation}

ProCLIP avoids the above issue through a progressive alignment strategy. Importantly, both the instance-level distillation loss $\mathcal{L}_{\text{ins}}$ and the structural distillation loss $\mathcal{L}_{\text{struct}}$ are used in both stages, but they play different roles at different stages.

In Stage I, the two distillation losses serve as inheritance constraints. The instance-level distillation loss provides point-wise semantic anchoring by pulling each projected LLM embedding $t_i^*$ toward its corresponding CLIP text embedding $e_i$. The structural distillation loss provides global geometric anchoring by preserving the pairwise semantic relations among CLIP text embeddings. Together, they calibrate the LLM representation space into a CLIP-compatible text manifold before cross-modal contrastive learning.

In Stage II, the same two losses serve as preservation constraints. While the contrastive loss refines the visual-language alignment, $\mathcal{L}_{\text{ins}}$ restricts instance-level representation drift and $\mathcal{L}_{\text{struct}}$ prevents the global semantic topology from being distorted. Therefore, instance distillation controls local semantic consistency, whereas structural distillation controls global geometric consistency throughout the entire progressive alignment process.

\subsection{Instance Distillation Provides Point-wise Semantic Anchoring}

We first analyze the role of instance-level distillation. In Stage I, $\mathcal{L}_{\text{ins}}$ directly constrains the projected LLM embedding to approach the corresponding CLIP text embedding. This provides a bounded initialization for the subsequent cross-modal contrastive learning.

\begin{theorem}[Bounded Initial Cross-Modal Discrepancy]
Assume that the embedding space is equipped with a metric distance $d(\cdot,\cdot)$. Let $\epsilon_C$ denote the original CLIP cross-modal alignment error, such that
\begin{equation}
    \mathbb{E}\left[d(v_i^{(0)}, e_i)\right] \leq \epsilon_C.
\end{equation}
If the instance-level distillation in Stage I learns an adapter such that
\begin{equation}
    \mathbb{E}\left[d(t_i^*, e_i)\right] \leq \epsilon_A,
\end{equation}
then the expected discrepancy between the original visual embedding and the projected LLM embedding before Stage II is bounded by
\begin{equation}
    \mathbb{E}\left[d(v_i^{(0)}, t_i^*)\right] \leq \epsilon_C + \epsilon_A.
\end{equation}
\end{theorem}

\begin{proof}
By the triangle inequality, for each image-text pair $(I_i,T_i)$, we have
\begin{equation}
    d(v_i^{(0)}, t_i^*) \leq d(v_i^{(0)}, e_i) + d(e_i, t_i^*).
\end{equation}
Taking the expectation over the data distribution gives
\begin{equation}
    \mathbb{E}\left[d(v_i^{(0)}, t_i^*)\right]
    \leq
    \mathbb{E}\left[d(v_i^{(0)}, e_i)\right]
    +
    \mathbb{E}\left[d(e_i, t_i^*)\right]
    \leq
    \epsilon_C + \epsilon_A.
\end{equation}
\end{proof}

This result shows that instance-level distillation provides an explicitly bounded initialization for Stage II. In contrast, without such point-wise anchoring, the distance between the projected LLM embedding and the CLIP visual embedding is not controlled by the original CLIP alignment structure and can be as large as the diameter of the representation space. Therefore, $\mathcal{L}_{\text{ins}}$ reduces the initial cross-modal gap and stabilizes the subsequent contrastive optimization.

In Stage II, $\mathcal{L}_{\text{ins}}$ further acts as an instance-level preservation constraint. It prevents each trainable representation from drifting excessively away from its CLIP-compatible reference. Thus, the role of instance distillation is two-fold: it first anchors the projected LLM embeddings to the CLIP text space, and then preserves local semantic consistency during contrastive refinement.

\subsection{Structural Distillation Preserves Semantic Topology}

Instance-level distillation aligns each projected LLM embedding with its corresponding CLIP text embedding. However, point-wise matching alone does not explicitly ensure that the relative semantic structure among different samples is preserved. To address this, ProCLIP introduces a structural distillation loss that encourages the projected LLM embeddings to inherit the pairwise geometry of the original CLIP text embeddings.

Let $\mathcal{E}=\{e_i\}_{i=1}^{N}$ denote the empirical CLIP text embedding set, and $\mathcal{T}^{*}=\{t_i^*\}_{i=1}^{N}$ denote the corresponding projected LLM embedding set. The structural distillation objective can be formulated as
\begin{equation}
    \mathcal{L}_{\text{struct}}
    =
    \sum_{i<j}
    \left|
    d(t_i^*,t_j^*) - d(e_i,e_j)
    \right|.
\end{equation}

\begin{proposition}[Empirical Semantic Topology Preservation]
If the structural alignment error satisfies
\begin{equation}
    \left|
    d(t_i^*,t_j^*) - d(e_i,e_j)
    \right|
    \leq \delta,
    \quad \forall i,j,
\end{equation}
then the projected LLM embedding set $\mathcal{T}^{*}$ forms a $\delta$-approximate isometry of the empirical CLIP text embedding set $\mathcal{E}$.
\end{proposition}

\begin{proof}
The definition of additive approximate isometry requires that pairwise distances are preserved up to an additive error $\delta$. Since for all pairs $(i,j)$,
\begin{equation}
    \left|
    d(t_i^*,t_j^*) - d(e_i,e_j)
    \right|
    \leq \delta,
\end{equation}
the pairwise geometry of $\mathcal{T}^{*}$ approximates that of $\mathcal{E}$ within error $\delta$. Therefore, the projected LLM embeddings preserve the empirical semantic topology of the original CLIP text space.
\end{proof}

This structural constraint is important because CLIP's zero-shot ability relies not only on image-text instance matching, but also on the global organization of semantic concepts. For example, semantically related concepts such as ``dog'' and ``cat'' should remain closer than unrelated concepts such as ``dog'' and ``airplane''. By preserving the pairwise structure of the original CLIP text space, $\mathcal{L}_{\text{struct}}$ prevents the adapter from learning only isolated point-wise correspondences and encourages the projected LLM space to inherit the global semantic geometry of CLIP.

In Stage II, structural distillation further prevents topology distortion during visual-language contrastive tuning. Even when the contrastive loss updates the representations to improve image-text matching, $\mathcal{L}_{\text{struct}}$ constrains the relative organization among samples. Therefore, the model can acquire improved LLM-based language understanding while maintaining the semantic structure of the original CLIP manifold.

\subsection{Why Short-Text Distillation Benefits Long-Text Understanding}

A natural question is why distillation from short CLIP-style texts can improve the final model's ability to understand longer and richer descriptions. The key reason is that Stage I does not aim to teach the LLM new language knowledge. Instead, it learns a geometric interface that maps the LLM representation space into the CLIP-compatible embedding space.

Let $g_{\phi}(\cdot)$ denote the adapter that projects an LLM representation into the CLIP text space. The LLM encoder itself already contains rich compositional and contextual knowledge learned from large-scale language pre-training. For a short text $s_i$, Stage I minimizes the discrepancy between $g_{\phi}(\mathrm{LLM}(s_i))$ and the original CLIP text embedding $e_i$. This process teaches the adapter how to translate LLM representations into the CLIP-compatible coordinate system.

The important point is that the learned mapping $g_{\phi}$ is shared across inputs. Once it learns how to align the local semantic directions of the LLM space with those of the CLIP text space using short texts, the same mapping can be applied to longer texts. Long texts can be regarded as compositions of multiple semantic units, such as objects, attributes, relations, actions, and contextual descriptions. If the adapter preserves the semantic directions of these basic units, then the projected representation of a long text can still lie in a CLIP-compatible region.

We can illustrate this with a simple compositional view. Suppose a long-text representation can be approximately decomposed into several semantic components:
\begin{equation}
    h_{\text{long}} \approx \sum_{k=1}^{K} \alpha_k h_k,
\end{equation}
where $h_k$ denotes the LLM representation of a semantic unit and $\alpha_k$ denotes its contextual weight. If the adapter is locally smooth around these semantic components, then
\begin{equation}
    g_{\phi}(h_{\text{long}})
    \approx
    g_{\phi}\left(\sum_{k=1}^{K} \alpha_k h_k\right).
\end{equation}
For a linear adapter, this further becomes
\begin{equation}
    g_{\phi}(h_{\text{long}})
    \approx
    \sum_{k=1}^{K} \alpha_k g_{\phi}(h_k).
\end{equation}
Even for a nonlinear adapter, this approximation holds locally when the adapter is smooth. Therefore, short-text distillation can align the basic semantic directions, while the LLM provides the compositional capability needed for long-text understanding.

Instance-level and structural distillation play complementary roles in this transfer. The instance-level loss anchors individual short-text semantics to the CLIP text space, while the structural loss preserves the relative relations among semantic concepts. Since long texts are often formed by composing multiple concepts and relations, preserving both local semantic anchors and global semantic topology makes the projected long-text representation more likely to remain meaningful in the CLIP-compatible space.

This explains why short-text supervision does not limit ProCLIP to short-text semantics. Instead, it calibrates the LLM representation into the CLIP embedding space, allowing the model to exploit the LLM's inherent long-context and compositional understanding ability during inference. In other words, distillation learns the interface, while the LLM supplies the linguistic expressiveness.

\subsection{Why Multilingual Ability Emerges without Multilingual Image-Text Data}

Another important property of ProCLIP is that it can improve multilingual understanding even though the training process does not require multilingual image-text pairs. This can be explained by separating language knowledge from visual alignment.

The original CLIP text encoder is usually limited in its language coverage and often performs best on English prompts. In contrast, modern LLMs encode different languages into a relatively shared semantic space through large-scale multilingual or cross-lingual pre-training. For semantically equivalent texts in different languages, their LLM representations tend to be closer than those produced by a monolingual CLIP text encoder. Let $x^{\text{en}}$ and $x^{\text{ml}}$ denote an English text and its multilingual counterpart with similar meaning. The LLM encoder naturally encourages
\begin{equation}
    \mathrm{LLM}(x^{\text{en}}) \approx \mathrm{LLM}(x^{\text{ml}})
\end{equation}
in the semantic representation space.

During training, ProCLIP learns a language-agnostic adapter $g_{\phi}$ that maps LLM semantic representations into the CLIP-compatible space. Importantly, this adapter is not tied to a specific language tokenization pattern, but operates on the LLM representation. Therefore, if two texts in different languages are close in the LLM semantic space, their projected embeddings will also be close after the adapter, provided that $g_{\phi}$ is smooth:
\begin{equation}
    \left\|
    g_{\phi}(\mathrm{LLM}(x^{\text{en}}))
    -
    g_{\phi}(\mathrm{LLM}(x^{\text{ml}}))
    \right\|
    \leq
    L_g
    \left\|
    \mathrm{LLM}(x^{\text{en}})
    -
    \mathrm{LLM}(x^{\text{ml}})
    \right\|,
\end{equation}
where $L_g$ is the Lipschitz constant of the adapter.

This inequality shows that the cross-lingual semantic proximity encoded by the LLM can be preserved after projection. As a result, although ProCLIP may use only image-text alignment data in a single language, the learned adapter can transfer the visual alignment to other languages through the multilingual semantic space already embedded in the LLM.

The two distillation losses further support this transfer. Instance-level distillation maps the LLM semantic representation into the CLIP-compatible space, while structural distillation preserves the relative semantic geometry among concepts. If multilingual expressions with similar meanings are already close in the LLM space, then the learned adapter can project them into similar CLIP-compatible regions. Therefore, the multilingual ability of ProCLIP does not come from multilingual image-text supervision. Instead, it comes from the combination of two factors: 
(1) the LLM provides a multilingual semantic representation space, and 
(2) ProCLIP learns a CLIP-compatible projection from this semantic space to the visual-language space. 
In this sense, ProCLIP acts as a bridge that connects the multilingual knowledge of the LLM with the visual grounding ability of CLIP.

\subsection{Stage-II Distillation Mitigates Representation Drift}

During Stage II, ProCLIP optimizes the contrastive objective to refine cross-modal alignment. However, if this objective is used alone, both the visual representation and the projected LLM representation may drift away from the original CLIP semantic space. To prevent this, ProCLIP continues to use instance-level and structural distillation losses as preservation constraints.

Let $z_i$ denote a trainable representation during Stage II, and let $z_i^{(0)}$ denote the corresponding CLIP-compatible reference representation. The instance-level distillation loss penalizes local deviations:
\begin{equation}
    \mathcal{L}_{\text{ins}}
    =
    \sum_i
    d(z_i, z_i^{(0)}).
\end{equation}
This term prevents each individual sample from drifting excessively.

The structural distillation loss further preserves the global semantic geometry:
\begin{equation}
    \mathcal{L}_{\text{struct}}
    =
    \sum_{i<j}
    \left|
    d(z_i,z_j)-d(z_i^{(0)},z_j^{(0)})
    \right|.
\end{equation}
Unlike instance-level regularization, which constrains each point independently, structural distillation constrains the relative organization among samples. Therefore, even when contrastive learning updates the representation for better image-text matching, the global semantic topology inherited from CLIP is still preserved.

We further analyze the effect of instance-level drift regularization. For simplicity, consider the induced optimization in the visual representation space. Let $v_i^{(0)}$ denote the visual embedding extracted by the original frozen CLIP encoder, and let $v_i$ denote the visual embedding during training. The Stage-II objective for one instance can be written as
\begin{equation}
    \mathcal{L}_{\text{tune}}(v_i)
    =
    \mathcal{L}_{\text{info}}(v_i,t_i^*)
    +
    \lambda_{\text{ins}}
    \left\|
    v_i - v_i^{(0)}
    \right\|_2^2.
\end{equation}

\begin{theorem}[Bounded Tangential Representation Drift]
Assume that the embeddings are normalized and the temperature of InfoNCE is fixed. Then the gradient norm of the InfoNCE loss is bounded by a constant $G$. At a constrained stationary point of $\mathcal{L}_{\text{tune}}$ on the unit sphere, the tangential component of the instance-level representation drift is bounded by
\begin{equation}
    \left\|
    \Pi_{v_i}
    (v_i - v_i^{(0)})
    \right\|_2
    \leq
    \frac{G}{2\lambda_{\text{ins}}},
\end{equation}
where $\Pi_{v_i}=I-v_i v_i^\top$ projects a vector onto the tangent space at $v_i$.
\end{theorem}

\begin{proof}
Taking the Euclidean gradient of the objective with respect to $v_i$, we have
\begin{equation}
    \nabla_{v_i}\mathcal{L}_{\text{tune}}
    =
    \nabla_{v_i}\mathcal{L}_{\text{info}}
    +
    2\lambda_{\text{ins}}
    (v_i-v_i^{(0)}).
\end{equation}
Because $v_i$ is constrained to the unit sphere, stationarity should be evaluated in the tangent space rather than by setting the Euclidean gradient to zero. Thus, at a constrained stationary point,
\begin{equation}
    \Pi_{v_i}
    \nabla_{v_i}\mathcal{L}_{\text{tune}}
    =
    0.
\end{equation}
Therefore,
\begin{equation}
    2\lambda_{\text{ins}}
    \Pi_{v_i}
    (v_i-v_i^{(0)})
    =
    -
    \Pi_{v_i}
    \nabla_{v_i}\mathcal{L}_{\text{info}}.
\end{equation}
Taking the $L_2$ norm on both sides gives
\begin{equation}
    \left\|
    \Pi_{v_i}
    (v_i - v_i^{(0)})
    \right\|_2
    \leq
    \frac{1}{2\lambda_{\text{ins}}}
    \left\|
    \Pi_{v_i}
    \nabla_{v_i}\mathcal{L}_{\text{info}}
    \right\|_2.
\end{equation}
Since projection cannot increase the norm, and the gradient norm of InfoNCE is bounded by $G$ under normalized embeddings and fixed temperature, we obtain
\begin{equation}
    \left\|
    \Pi_{v_i}
    (v_i-v_i^{(0)})
    \right\|_2
    \leq
    \frac{G}{2\lambda_{\text{ins}}}.
\end{equation}
\end{proof}

This bound shows that $\lambda_{\text{ins}}$ explicitly controls the trade-off between learning new LLM-based alignment and preserving the original CLIP representation. A larger $\lambda_{\text{ins}}$ imposes a stronger constraint on local representation drift. Meanwhile, $\mathcal{L}_{\text{struct}}$ constrains the global pairwise geometry. Therefore, the two distillation losses jointly mitigate representation degradation from both local and global perspectives.

\subsection{Discussion: From Direct Alignment to Progressive Alignment}

The above analysis reveals the essential difference between direct alignment and ProCLIP. Direct alignment attempts to align the visual encoder with an uncalibrated LLM embedding space. This makes the visual encoder vulnerable to unstable contrastive gradients and may degrade its pre-trained semantic structure. In contrast, ProCLIP first aligns the LLM representation with the CLIP text manifold and then performs visual-language refinement under continuous distillation constraints.

This progressive design has three advantages. First, instance-level distillation provides point-wise semantic anchoring, reducing the initial cross-modal gap and preventing the optimization from starting from an uncalibrated LLM space. Second, structural distillation preserves the global semantic topology of CLIP, preventing the learned representation from collapsing into isolated instance-level matches. Third, because the adapter operates on LLM semantic representations rather than language-specific tokens, the long-text and multilingual knowledge already encoded in the LLM can be projected into the CLIP-compatible visual-language space.

Together, these mechanisms explain why ProCLIP can leverage the richer linguistic, compositional, and multilingual knowledge of LLMs while preserving the robust visual generalization ability.

\section{More results and analysis.}
\noindent\textbf{Detailed comparison on fine-grained and multilingual benchmarks.} Table~\ref{tab:detail_result1} compares the fine-grained and multilingual capabilities of different methods. S.C. in the table is the abbreviation for SugarCrepe. ProCLIP consistently surpasses LLM2CLIP across SugarCrepe, XM3600, and MMVP under different CLIP architectures and data scales. The gains on SugarCrepe demonstrate that ProCLIP better captures fine-grained compositional relations, including replacement, swapping, and addition. Meanwhile, its improvements on XM3600 I2T/T2T indicate stronger multilingual cross-modal alignment. ProCLIP also achieves notable gains on MMVP, suggesting enhanced visual discrimination in challenging fine-grained understanding tasks. These consistent improvements verify the effectiveness of ProCLIP in both compositional reasoning and multilingual generalization.

\noindent\textbf{Detailed comparison of linear probe performance.} Table~\ref{tab:linear} summarizes the detailed linear probe performance across 11 downstream classification datasets. ProCLIP consistently outperforms LLM2CLIP in average accuracy across different architectures and data scales, demonstrating stronger transferable visual representations. The improvements are particularly evident on fine-grained datasets such as Food101, Cars, Aircraft, DTD, Pets, and Flowers, indicating that ProCLIP better preserves discriminative visual features. Notably, ProCLIP also remains competitive with the original CLIP baseline, suggesting that it enhances language-guided adaptation while avoiding degradation of the general-purpose representation capability.

\input{Table/sugar_crepe_mmvp_xm3600}
\input{Table/liner_probe}

\noindent\textbf{Analysis of Data Scale and Model Scale.} We further analyze the effects of data scale and model scale, as shown in Tab.~\ref{tab:sacle_analysis}.
For data scale, we observe that model performance improves with increasing data size. For example, when trained on 3M samples, ProCLIP achieves a zero-shot IN-1k accuracy of 62.5\%, which rises to 67.8\% when the dataset size increases to 30M. Under the same data scale, ProCLIP consistently outperforms LLM2CLIP. Notably, when we randomly sample 1M images from CC3M for training, ProCLIP achieves comparable or even superior zero-shot retrieval performance compared with LLM2CLIP, reaching 61.8\% on zero-shot IN-1k. This highlights the data efficiency of ProCLIP.
For model scale, we further expand the linear layers by three times, using $12\times$ layers in total, which leads to additional performance gains. This suggests that ProCLIP can continue to benefit from simple parameter scaling.
\input{Table/scale_analysis}

\noindent\textbf{Detailed comparison of robustness performance.} Table~\ref{tab:robustness} compares the robustness performance on ImageNet distribution-shift benchmarks. ProCLIP consistently surpasses LLM2CLIP across all CLIP architectures and training data scales, showing stronger generalization under distribution shifts. The gains are particularly notable on IN-A, IN-R, and IN-S, where robust visual recognition is more challenging. Compared with LLM2CLIP, ProCLIP better preserves the pre-trained robustness of CLIP while improving adaptation effectiveness. These results demonstrate that ProCLIP can effectively mitigate robustness degradation and maintain more reliable visual representations under shifted distributions.

\noindent\textbf{Generalized to pixel-level tasks.} Table~\ref{tab:ovss} reports the open-vocabulary semantic segmentation performance on ADE20K, COCO-Stuff164k, and VOC20. Compared with the original CLIP, LLM2CLIP brings moderate improvements on all three datasets, indicating that language-side enhancement can benefit dense prediction transfer to some extent. In contrast, ProCLIP achieves the best results across all benchmarks, improving over LLM2CLIP\input{Table/robutness_v2} by 1.4, 1.8, and 1.2 points on ADE20K, COCO-Stuff164k, and VOC20, respectively. These consistent gains suggest that ProCLIP better preserves and transfers CLIP's visual-language alignment to pixel-level recognition, leading to stronger open-vocabulary segmentation capability.

\noindent\textbf{Efficiency analysis.}
Table 23 compares the training efficiency of ProCLIP and LLM2CLIP under the same $8\times$H100 GPU setting. ProCLIP requires slightly higher VRAM usage and training time than LLM2CLIP, increasing from 32.6 GB to 43.2 GB and from 1.2 to 1.7 hours, respectively. However, the overall overhead is moderate, suggesting that ProCLIP maintains good training efficiency while achieving stronger performance.

\input{Table/ov_ss}
\input{Table/efficiency_llm2clip_proclip}

\section{Visualization.}
\begin{figure}[t!]
    \centering
    \includegraphics[width=0.9\textwidth]{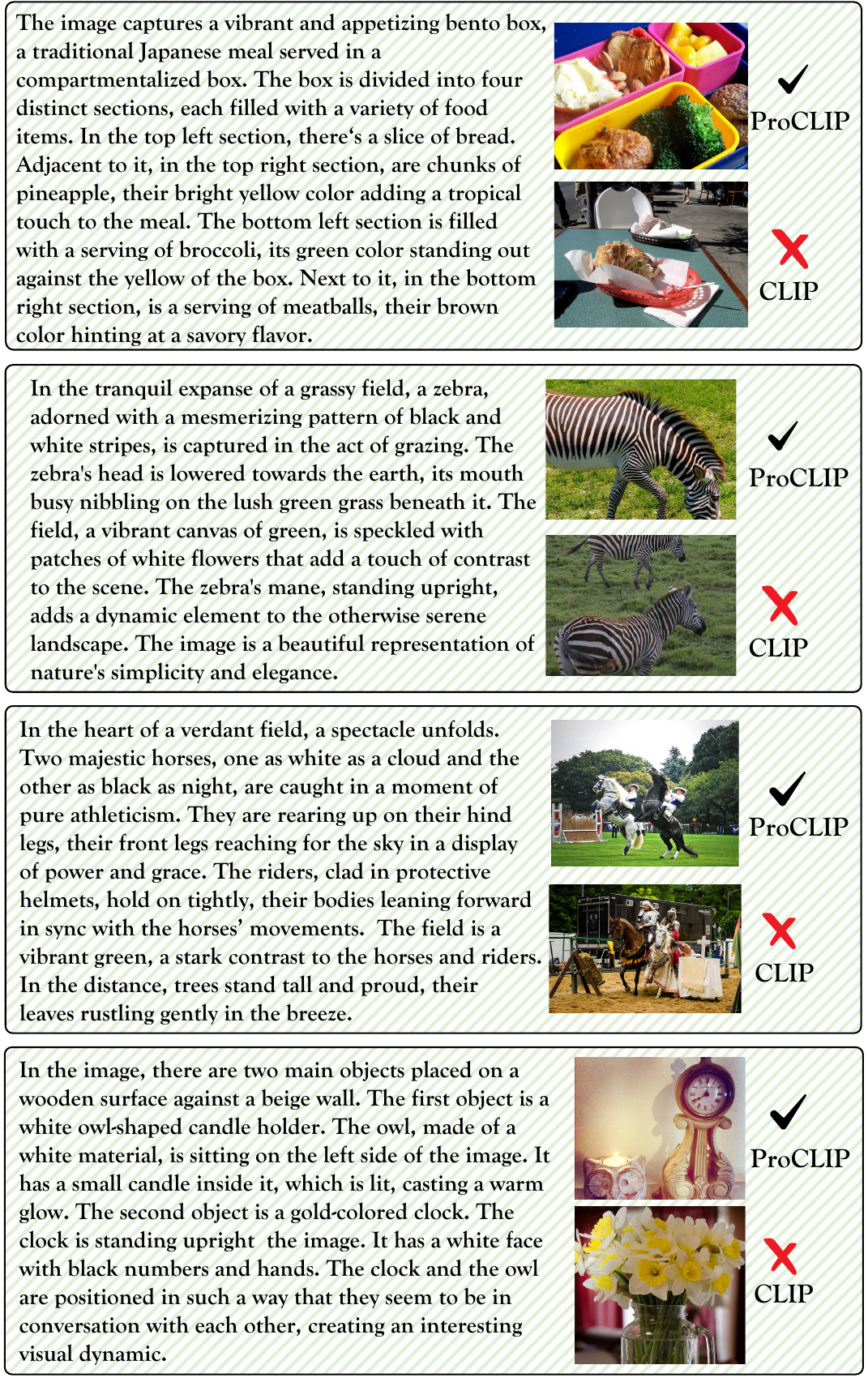}

    \caption{The visualization of ProCLIP vs CLIP in long-text cross-modal retrieval. From top to bottom: Case 1, Case 2, Case 3, and Case 4.}
    \label{fig:four_cases_vertical}
\end{figure}

As illustrated in Fig.~\ref{fig:four_cases_vertical}, ProCLIP effectively extends long-text processing capabilities, enabling precise image-text alignment even within highly detailed descriptions. Moreover, ProCLIP extends CLIP's capabilities to include multilingual support and enhanced fine-grained understanding, which are absent in the original model.
\section{Future Works.}

\textbf{Training Efficiency.} ProCLIP is subject to the limitation of increased computational overhead. We consider the following directions to potentially reduce computational overhead:
\begin{itemize}
    \item Adopting a PEFT-based approach to fine-tune the vision encoder in the second stage
    \item 
Fine-tuning only part of the vision encoder parameters in the second stage, such as the last few Transformer blocks
\item 
Replacing online distillation with offline distillation, which would substantially reduce the additional computational cost introduced in the second stage
\end{itemize}

\noindent\textbf{Fine-grained Visual-Text Alignment.} 
Future work explores the integration of a dedicated local alignment loss to explicitly strengthen the model’s fine-grained visual-textual alignment capabilities, thereby enhancing performance on tasks that require detailed spatial understanding.

\noindent\textbf{More Model Architecture.} Our approach replaces the original CLIP text encoder with an LLM-based embedder to enhance multiple capabilities. From another perspective, we consider whether the vision encoder in the dual-tower architecture can also be replaced to address limitations in visual representation.

%% file: Table/sugar_crepe_tongji.tex
\begin{table*}[t!]
\centering
\caption{Number of hard negative captions of all types in SugarCrepe.}

\begin{tabular}{lccccccc} %
\toprule
 & \multicolumn{3}{c}{\textbf{REPLACE}} & \multicolumn{2}{c}{\textbf{SWAP}} & \multicolumn{2}{c}{\textbf{ADD}} \\
\cmidrule(lr){2-4} \cmidrule(lr){5-6} \cmidrule(lr){7-8}
 & Object & Attribute & Relation & Object & Attribute & Object & Attribute \\
\midrule
\# negative captions & 1652 & 788 & 1406 & 246 & 666 & 2062 & 692 \\
\bottomrule
\end{tabular}

\label{tab:negative_captions}

\end{table*}

%% file: Table/sugar_crepe_mmvp_xm3600.tex
\begin{table}[t!]
    \centering
\caption{Detailed comparison of fine-grained and multilingual capability on SugarCrepe, XM3600, and MMVP across different CLIP architectures and data scales.}
    \resizebox{\textwidth}{!}{
    \begin{tabular}{lccccccc}
    \toprule
       \textbf{ Method} & \textbf{\#Data} & \textbf{S.C. Replace} & \textbf{S.C. Swap} & \textbf{S.C. Add} & \textbf{XM3600 I2T} & \textbf{XM3600 T2T} & \textbf{MMVP}  \\
        \midrule
        \multicolumn{6}{l}{\textit{Model Architecture: CLIP ViT-B/32}}  \\
        CLIP & 400M & 80.2 & 64.0 & 72.7 & - & - & 15.6 \\
        \hdashline
        LLM2CLIP &3M  &87.3 & 72.5 & 85.1 & 40.3& 39.9 & 21.5\\
        \rowcolor[rgb]{0.925,0.957,1}
        ProCLIP & 3M & \textbf{88.0} &\textbf{72.9} & \textbf{87.5}&\textbf{43.0} & \textbf{43.8 }& \textbf{25.9}\\
          \hdashline
        LLM2CLIP &15M  &88.4 & 74.7 & 86.1 & 42.1 & 40.6  & \textbf{25.9}\\
        \rowcolor[rgb]{0.925,0.957,1}
        ProCLIP & 15M &\textbf{89.0} & \textbf{75.0} & \textbf{87.9} & \textbf{43.7} & \textbf{44.8} & 24.4\\
          \hdashline
        LLM2CLIP &30M  &88.3 &\textbf{75.4} & 85.6 & 43.9 & 41.8  & 25.2\\
        \rowcolor[rgb]{0.925,0.957,1}
        ProCLIP & 30M &\textbf{89.1} & 75.3 & \textbf{88.0} &\textbf{44.5}  & \textbf{45.2} & \textbf{30.4}\\

\midrule
        \multicolumn{6}{l}{\textit{Model Architecture: CLIP ViT-B/16}}  \\
        CLIP & 400M & 81.6 & 65.2 & 82.5& - & - & 12.6\\
          \hdashline
        LLM2CLIP &3M  &88.6 & 71.4 & 87.9 & 43.1 & 43.7&20.7 \\
        \rowcolor[rgb]{0.925,0.957,1}
        ProCLIP & 3M & \textbf{89.3} & \textbf{73.4} & \textbf{89.0} & \textbf{44.0} & \textbf{45.6}& \textbf{29.6}\\
          \hdashline
        LLM2CLIP &15M  & 89.3 & 75.1 & 89.3 & 44.3&  44.9 & 22.2\\
        \rowcolor[rgb]{0.925,0.957,1}
        ProCLIP & 15M &\textbf{89.7} & \textbf{75.9} & \textbf{89.9} & \textbf{45.0} & \textbf{45.6} & \textbf{24.4}\\
          \hdashline
        LLM2CLIP &30M  &89.4 & 73.9 & 89.5 & 44.7& 45.9 & 28.2\\
        \rowcolor[rgb]{0.925,0.957,1}
        ProCLIP & 30M &\textbf{89.5} & \textbf{74.8} & \textbf{91.0} &\textbf{45.9} &\textbf{46.3}  & \textbf{30.4}\\
\midrule
        \multicolumn{6}{l}{\textit{Model Architecture: CLIP ViT-L/14}}  \\
        CLIP & 400M & 79.5 & 61.3 & 74.9 & - & - & 19.3\\
          \hdashline
        LLM2CLIP &3M  &90.0 & 73.7 & 89.9 & 44.2 & 46.6 & 22.2\\
        \rowcolor[rgb]{0.925,0.957,1}
        ProCLIP & 3M &\textbf{90.7} &\textbf{75.5} &\textbf{ 92.5} & \textbf{46.1}&\textbf{49.1} & \textbf{25.2} \\
          \hdashline
        LLM2CLIP &15M &90.6 & 75.4 & 91.9 &44.8 & 47.8 & 25.2\\
        \rowcolor[rgb]{0.925,0.957,1}
        ProCLIP & 15M &\textbf{91.9} & \textbf{77.2} & \textbf{93.1} & \textbf{46.7} &\textbf{51.0}& \textbf{27.4}\\
          \hdashline
        LLM2CLIP &30M  &89.7 & 74.2 & 91.5  & 46.2 &  48.3 & 23.7\\
        \rowcolor[rgb]{0.925,0.957,1}
        ProCLIP & 30M &\textbf{90.3} &\textbf{75.4} & \textbf{92.8} &\textbf{ 47.8}& \textbf{52.6} & \textbf{34.1}\\
        \bottomrule
    \end{tabular}
    }
    \label{tab:detail_result1}
\end{table}

%% file: Table/liner_probe.tex
\begin{table*}[t!]
\centering
\captionsetup{font=small}
\caption{Detailed linear probe performance on 11 downstream classification datasets across different CLIP architectures and training data scales.}
\label{tab:linear}
\resizebox{\textwidth}{!}{
\begin{tabular}{lccccccccccccc}
\toprule
\textbf{Method}  &\textbf{\#Data} &  \rotatebox{90} {\textbf{Food101}} & \rotatebox{90} {\textbf{CIFAR10}} &  \rotatebox{90} {\textbf{CIFAR100}}& \rotatebox{90} {\textbf{SUN397}} & \rotatebox{90} {\textbf{Cars}} & \rotatebox{90} {\textbf{Aircraft}}& \rotatebox{90} {\textbf{DTD}} & \rotatebox{90} {\textbf{Pets}}  & \rotatebox{90} {\textbf{Caltech101}} &  \rotatebox{90} {\textbf{Flowers}}  & \rotatebox{90} {\textbf{ImageNet}} & \textbf{Avg.} \\
\midrule
\multicolumn{6}{l}{\textit{Model Architecture: CLIP ViT-B/32}} & & & &  & & & & \\
CLIP & 
400M & 
88.6 & 
95.1 & 
80.1 & 
73.4 & 
80.8 & 
44.9 & 
76.3 & 
89.3 & 
92.7 & 
94.7 & 
74.3 & 
80.9
\\
\hdashline
LLM2CLIP & 3M & 87.9 & 95.7 & \textbf{83.1} & 74.1 & 78.0 & \textbf{44.9} & 77.7 & \textbf{90.4} & 92.4 & 94.6 & 74.2 & 81.2 \\
\rowcolor[rgb]{0.925,0.957,1}\modelnm\ & 3M & \textbf{88.4} & \textbf{95.9} & \textbf{83.1} & \textbf{74.3} & \textbf{79.5} & 44.1 & \textbf{78.2} & 90.3 & \textbf{92.6} & \textbf{95.0} & \textbf{74.4} & \textbf{81.4}\\
\hdashline
LLM2CLIP & 15M & 87.7 & 95.7 & 82.7 & 74.0 & 77.5 & 44.2 & \textbf{78.3} & \textbf{90.2} & 92.5 & 94.4 &74.2 &81.0 \\
\rowcolor[rgb]{0.925,0.957,1}\modelnm\ & 15M & \textbf{88.7} & \textbf{95.9} & \textbf{82.8} & \textbf{74.8} & \textbf{80.8} & \textbf{44.9} & 78.1 & \textbf{90.2} & \textbf{92.8} & \textbf{95.1} & \textbf{74.4} & \textbf{81.7} \\
\hdashline
LLM2CLIP & 30M & 87.6 & 95.9 & 83.0 & 74.1 & 76.3 & 43.5 & 77.6 & \textbf{90.1} & \textbf{92.8} & 93.8 & 74.3& 80.8\\
\rowcolor[rgb]{0.925,0.957,1}\modelnm\ & 30M & \textbf{88.2} & \textbf{96.0} & \textbf{83.1} & \textbf{75.1} & \textbf{79.0} & \textbf{43.8} & \textbf{77.8} & 89.8 & 92.6 & \textbf{94.9} & \textbf{74.5} & \textbf{81.4}\\
\midrule
\multicolumn{6}{l}{\textit{Model Architecture: CLIP ViT-B/16}} & & & &  & & & & \\
CLIP & 
400M & 
92.7 & 
96.0 & 
82.5 & 
75.7 & 
85.9 & 
52.8 & 
78.9 & 
93.1 & 
93.9 & 
96.4 & 
79.6 & 
84.4\\

\hdashline
LLM2CLIP & 3M & 91.6 & \textbf{97.0} & 84.5 & 76.0 & 82.1 & 50.1 & 80.3 & 92.3 & 93.6 & 95.7 & 79.6& 83.9 \\
\rowcolor[rgb]{0.925,0.957,1}\modelnm\ & 3M & \textbf{92.8} & 96.8 & \textbf{84.6} & \textbf{76.4} & \textbf{85.6} & \textbf{52.0} & \textbf{80.6} & \textbf{93.3} & \textbf{94.2} & \textbf{97.0} &  \textbf{79.7} & \textbf{84.8} \\
\hdashline
LLM2CLIP & 15M & 91.9 & \textbf{97.0} & \textbf{84.9} & 75.6 & 83.7 & 50.7 & 80.4 & 92.9 & 93.8 & 96.6 &79.6 & 84.3\\
\rowcolor[rgb]{0.925,0.957,1}\modelnm\ & 15M & \textbf{92.6} & 96.7 & 84.3 & \textbf{76.6} & \textbf{85.6} & \textbf{51.4} & \textbf{80.8} & \textbf{93.6} & \textbf{94.3} & \textbf{96.7} & \textbf{79.8} & \textbf{84.8}\\
\hdashline
LLM2CLIP & 30M & 91.3 & \textbf{96.6} & 84.8 & 75.3 & 80.6 & 48.2 & 80.3 & 92.5 & 93.4 & 95.0 & 79.7& 83.4  \\
\rowcolor[rgb]{0.925,0.957,1}\modelnm\ & 30M & \textbf{92.3} & \textbf{96.6} & \textbf{85.7} & \textbf{77.0} & \textbf{84.7} & \textbf{50.1} & \textbf{81.2} & \textbf{93.1} & \textbf{94.0} & \textbf{96.7} & \textbf{79.5} & \textbf{84.6} \\
\midrule

\multicolumn{6}{l}{\textit{Model Architecture: CLIP ViT-L/14}} & & & &  & & & & \\
CLIP & 
400M & 
95.3 & 
89.1 & 
87.2 & 
79.4 & 
90.7 & 
63.0 & 
81.8 & 
95.3 & 
96.9 & 
98.8 & 
82.9 & 
88.1 \\
\hdashline
LLM2CLIP & 3M & 94.5  & \textbf{98.6} & \textbf{89.2} & 79.6 & 86.7 & 57.7 & 83.4 & 94.1 & 96.4 & 97.1 & \textbf{82.5} & 87.2 \\
\rowcolor[rgb]{0.925,0.957,1}\modelnm\ & 3M & \textbf{95.3} & 98.5 & 88.8 & \textbf{80.3} & \textbf{90.3} & \textbf{61.0} & \textbf{83.6} & \textbf{95.2} & \textbf{96.9} & \textbf{98.7} & 81.9 & \textbf{88.2} \\
\hdashline
LLM2CLIP & 15M & 94.4 & \textbf{98.5} & \textbf{88.8} & 78.5 & 86.0 & 55.0 & 82.7 & 93.9 & 95.9 & 97.1 & 82.6 & 86.7 \\
\rowcolor[rgb]{0.925,0.957,1}\modelnm & 15M & \textbf{95.2} & 98.4 & 88.6 & \textbf{79.7} & \textbf{90.5} & \textbf{61.4} & \textbf{83.3} & \textbf{95.3} & \textbf{96.8} & \textbf{98.7} & \textbf{83.0} & \textbf{86.7} \\
\hdashline
LLM2CLIP & 30M & 94.1 &  98.2& 88.4& 78.7& 84.6& 54.8 & 82.4 & 93.7 & 95.8 & 96.5 & 82.2 & 86.3 \\
\rowcolor[rgb]{0.925,0.957,1}\modelnm\ & 30M & \textbf{95.1} & \textbf{98.4} & \textbf{89.0} & \textbf{80.3} & \textbf{90.0} & \textbf{60.0} & \textbf{83.9} & \textbf{95.2} & \textbf{96.8} & \textbf{98.5} & \textbf{82.7} & \textbf{88.2} \\
\bottomrule
\end{tabular}
}
\vspace{-2mm}
\end{table*}

%% file: Table/scale_analysis.tex
\begin{table*}[t!]
    \centering

    \caption{Comparison of data and model scales under ViT-L architecture.}
    \begin{tabular}{cccccc}
    \toprule
        \textbf{Method} & \textbf{\#Data} & \textbf{Adapter} & \textbf{IN-1k} & \textbf{I2T Avg.} & \textbf{T2I Avg.}\\
        \hline
    \textcolor{gray}{CLIP} & \textcolor{gray}{400M} & \textcolor{gray}{-} &
    \textcolor{gray}{74.5} & \textcolor{gray}{66.9} & \textcolor{gray}{55.5}\\
        \hdashline
        \rowcolor[rgb]{0.925,0.957,1}
        ProCLIP & 1M & $4\times$linear & 61.8 & 81.9 & 79.7\\
        \hdashline
        LLM2CLIP & 3M & $4\times$linear & 52.8 & 82.0 & 79.1\\
        \rowcolor[rgb]{0.925,0.957,1}
        ProCLIP & 3M & $4\times$linear & \textbf{62.5} & \textbf{83.0} & \textbf{80.5}\\
        \hdashline
        LLM2CLIP & 15M & $4\times$linear & 54.0 & 83.2 & 80.1\\
        \rowcolor[rgb]{0.925,0.957,1}
        ProCLIP & 15M & $4\times$linear & \textbf{66.0} & \textbf{84.3} & \textbf{81.2} \\
        \hdashline
        LLM2CLIP & 30M & $4\times$linear & 56.4 & 85.8 & 82.1 \\
        \rowcolor[rgb]{0.925,0.957,1}
        ProCLIP & 30M &  $4\times$linear & 67.8 & 86.2 & 82.6 \\
        \rowcolor[rgb]{0.925,0.957,1}
        ProCLIP & 30M & $12\times$linear  & \textbf{71.5} & \textbf{86.8} & \textbf{82.8}\\
        \bottomrule
    \end{tabular}
    \vspace{-7mm}
    \label{tab:sacle_analysis}
\end{table*}

%% file: Table/robutness_v2.tex
\begin{wraptable}{r}{0.5\textwidth}
\centering
\caption{Detailed comparison of robustness performance on ImageNet distribution-shift benchmarks.}
\label{tab:robustness}
\resizebox{0.5\textwidth}{!}{
\begin{tabular}{lcccccc}
\toprule
\multirow{2}{*}{\textbf{Method}} 
& \multirow{2}{*}{\textbf{\#Data}} 
& \multicolumn{5}{c}{\textbf{Robustness}} \\
& & \textbf{IN-V2} & \textbf{IN-A} & \textbf{IN-O} & \textbf{IN-R} & \textbf{IN-S} \\
\midrule

\multicolumn{7}{l}{\textit{Model Architecture: CLIP ViT-B/32}} \\
CLIP & 400M & 55.9 & 31.6 & 47.8 & 69.3 & 42.3 \\
\hdashline
LLM2CLIP & 3M & 37.0 & 18.8 & 38.0 & 54.0 & 30.7 \\
\rowcolor[rgb]{0.925,0.957,1}
ProCLIP & 3M & \textbf{43.6} & \textbf{25.5} & \textbf{40.6} & \textbf{62.9} & \textbf{35.3} \\
LLM2CLIP & 15M & 38.8 & 20.1 & 40.5 & 58.3 & 33.4 \\
\rowcolor[rgb]{0.925,0.957,1}
ProCLIP & 15M & \textbf{48.3} & \textbf{28.3} & \textbf{44.0} & \textbf{67.4} & \textbf{38.9} \\
LLM2CLIP & 30M & 41.0 & 22.9 & 41.8 & 59.5 & 34.0 \\
\rowcolor[rgb]{0.925,0.957,1}
ProCLIP & 30M & \textbf{48.9} & \textbf{29.5} & \textbf{46.3} & \textbf{67.8} & \textbf{39.0} \\

\midrule
\multicolumn{7}{l}{\textit{Model Architecture: CLIP ViT-B/16}} \\
CLIP & 400M & 61.9 & 49.9 & 42.3 & 77.7 & 48.2 \\
\hdashline
LLM2CLIP & 3M & 42.4 & 30.2 & 35.3 & 61.6 & 35.2 \\
\rowcolor[rgb]{0.925,0.957,1}
ProCLIP & 3M & \textbf{50.8} & \textbf{41.4} & \textbf{38.5} & \textbf{71.3} & \textbf{41.7} \\
LLM2CLIP & 15M & 44.8 & 33.9 & 37.3 & 65.4 & 38.1 \\
\rowcolor[rgb]{0.925,0.957,1}
ProCLIP & 15M & \textbf{54.8} & \textbf{45.3} & \textbf{40.5} & \textbf{76.0} & \textbf{45.2} \\
LLM2CLIP & 30M & 46.0 & 35.6 & 38.3 & 67.0 & 38.7 \\
\rowcolor[rgb]{0.925,0.957,1}
ProCLIP & 30M & \textbf{55.2} & \textbf{46.0} & \textbf{41.9} & \textbf{75.9} & \textbf{44.9} \\

\midrule
\multicolumn{7}{l}{\textit{Model Architecture: CLIP ViT-L/14}} \\
CLIP & 400M & 69.8 & 70.8 & 32.2 & 87.8 & 59.6 \\
\hdashline
LLM2CLIP & 3M & 49.0 & 46.6 & 32.4 & 75.0 & 44.8 \\
\rowcolor[rgb]{0.925,0.957,1}
ProCLIP & 3M & \textbf{58.3} & \textbf{63.3} & \textbf{31.6} & \textbf{84.0} & \textbf{52.3} \\
LLM2CLIP & 15M & 50.8 & 50.1 & 33.8 & 78.2 & 46.3 \\
\rowcolor[rgb]{0.925,0.957,1}
ProCLIP & 15M & \textbf{62.1} & \textbf{66.4} & \textbf{34.2} & \textbf{86.4} & \textbf{55.3} \\
LLM2CLIP & 30M & 52.7 & 52.7 & 34.0 & 78.6 & 47.3 \\
\rowcolor[rgb]{0.925,0.957,1}
ProCLIP & 30M & \textbf{63.4} & \textbf{68.0} & \textbf{34.1} & \textbf{86.8} & \textbf{55.7} \\

\bottomrule
\end{tabular}
}
\end{wraptable}

%% file: Table/ov_ss.tex
\begin{table}[t!]
    \centering
\caption{Comparison of open-vocabulary semantic segmentation performance. }
    \begin{tabular}{lcccc}
    \toprule
         \textbf{Method} &  \textbf{\#Data}  & \textbf{ADE20K}  & \textbf{COCO-Stuff164k} & \textbf{VOC20} \\
         \midrule
         \multicolumn{3}{l}{\textit{Model Architecture: CLIP ViT-L/14}} \\
         CLIP & 400M & 1.4 & 2.8 & 16.3\\
         LLM2CLIP  &30M & 2.4 & 3.0 & 17.5 \\
         \rowcolor[rgb]{0.925,0.957,1}
         ProCLIP &30M & \textbf{3.8} & \textbf{4.8}& \textbf{18.7} \\
         \bottomrule
    
    \end{tabular}
    \label{tab:ovss}
\end{table}

%% file: Table/efficiency_llm2clip_proclip.tex
\begin{table}[t!]
\centering
\caption{Training efficiency comparison between ProCLIP and LLM2CLIP.}
\label{tab:efficiency_llm2clip_proclip}

\begin{tabular}{lccc}
\toprule
\textbf{Method} & \textbf{Data} & \textbf{VRAM Usage (GB) $\downarrow$} & \textbf{Training Hours $\downarrow$} \\
\midrule
\multicolumn{3}{l}{\textit{Model Architecture: CLIP ViT-L/14}}  \\
LLM2CLIP & 3M & \textbf{32.6} & \textbf{1.2} \\
\rowcolor[rgb]{0.925,0.957,1}
ProCLIP    & 3M  & 43.2  & 1.7\\
\bottomrule
\end{tabular}

\end{table}